\documentclass[11pt]{article}

\usepackage[margin=1in]{geometry}
\usepackage{graphicx}
\usepackage{pdflscape}
\usepackage{xcolor}
\usepackage{amsmath,amssymb}
\usepackage{graphicx}
\usepackage[T1]{fontenc}
\usepackage[utf8]{inputenc}
\usepackage{lmodern}
\usepackage{microtype}
\usepackage{booktabs}
\usepackage{longtable}
\usepackage{array}
\usepackage[rightcaption]{sidecap}
\usepackage{caption}
\usepackage{microtype}
\usepackage{lineno}
\usepackage{xurl}

\usepackage{silence}
\usepackage{setspace}
\usepackage[numbers,sort&compress]{natbib}
\usepackage{hyperref}
\hypersetup{
  pdftitle={Integrating national forest inventory, airborne lidar, and satellite imagery for wall-to-wall mapping of forest structure with computer vision},
  pdfauthor={Vibrant Planet},
  hidelinks,
  colorlinks=true,
  linkcolor=blue,
  citecolor=blue,
  urlcolor=blue
}

\usepackage{placeins} % provides \FloatBarrier

\usepackage{fancyhdr}
\fancypagestyle{firstpage}{
  \fancyhf{}
  \fancyfoot[L]{\footnotesize\rule{0.35\textwidth}{0.4pt}\\[0.3em]
  $^{*}$Corresponding author: \textbf{ddiaz@vibrantplanet.net}\\
  $^{\dagger}$Work carried out while at Vibrant Planet.}

}

\title{
Integrating national forest inventory, airborne lidar,\\and satellite 
imagery for wall-to-wall mapping\\of forest structure with computer vision
}

\author{
Luke J. Zachmann, David D. Diaz$^{*}$, Vincent A. Landau, Chelsey Walden-Schreiner,\\
Tony Chang, Nathan E. Rutenbeck, Katharyn A. Duffy, Kiarie Ndegwa, \\
Andreas Gros$^{\dagger}$, Scott Conway, Guy Bayes$^{\dagger}$\\[1em]
Vibrant Planet Public Benefit Corporation, Truckee, CA, USA
}

\date{April 2026}

\begin{document}
% \linenumbers
\maketitle
\thispagestyle{firstpage}

\begin{abstract}
Remote sensing is increasingly relied upon to deliver actionable science for forest and wildfire risk management across large landscapes. Wall-to-wall, annually updated maps are a persistent need for effective forest management. Many planning systems and data collections combine disparate data sources with different purposes, vintages, and prediction quality, which leads to confounding behavior in operational planning systems. We introduce the VibrantForests framework, developed and applied to map forest attributes and provide a coherent foundation for effective forest and wildfire planning. VibrantForests includes a satellite-based forest structure model trained on lidar-derived samples and applied across the contiguous United States to concurrently generate estimates of canopy cover, canopy height, aboveground live tree biomass, basal area, and quadratic mean  diameter at 10-meter resolution. We demonstrate predictive capability spanning the full spectrum of forest conditions ranging from sparse-canopy/low-biomass to dense-canopy/high-biomass. Results show that our model extends the range at which saturation is commonly encountered in comparable passive-sensor models, and reduces regression-to-mean behavior that commonly produces overestimation of forest attributes in small/sparse conditions and underestimation in large/dense conditions. The VibrantForests framework addresses a key limitation in large-area forest and wildfire planning by delivering coherent wall-to-wall estimates of management-relevant attributes at annual cadence and 10m resolution.

\end{abstract}

\section{Introduction}\label{sec:intro}
Forest owners, managers, and decision-makers involved in the stewardship of forest landscapes and communities around them require a broad set of metrics to effectively inform land management decisions and to monitor outcomes. The growing demand among forest managers for actionable data at scale is only tangentially addressed by significant recent private-sector investment and model-development efforts for global-scale remote sensing applications that predict a narrower set of forest attributes related to carbon storage and sequestration \cite{anderson-2026, tolan-2024, liu-2023, -p}. 

Over several years,  Vibrant Planet has developed and implemented a decision support system for the assessment of wildfire risk and the land management planning \cite{safford-2026}. The Vibrant Planet Platform is now in use across more than 177 million acres in the United States. Developers and early users of the Vibrant Planet Platform commonly identified a variety of internal discrepancies and unexpected behaviors that emerged from mixing and combining datasets that had been designed by different data providers, for disparate purposes, and with inconsistent vintages and update schedules. These inconsistencies combined to produce unexpected artifacts in the data and downstream dependencies like forest treatment recommendations. To address these issues, we determined a more coherent modeling system was needed.

The Vibrant Planet Platform operates at spatial scales ranging from individual forest stands to landscapes spanning hundreds or thousands of watersheds. With this in mind, we identified several core requirements for a forest structure modeling system for these use cases, including: an annual or more frequent update cadence; spatial resolution no coarser than 10-30 meters; a consistent, low-cost input data stream; and extensibility to geographies beyond the contiguous United States. We determined that publicly available satellite imagery was the optimal choice for raw input data to meet these requirements. 

Vibrant Planet thus undertook an effort to construct a coherent modeling framework, based primarily on globally available satellite imagery. The intended user of these data products are natural resource managers and other professionals leading wildfire risk assessments, community wildfire protection planning, and forest restoration planning. These use cases generally involve access to summaries of forest and other land attributes at the scale of forest stands or treatment areas (polygons encompassing approximately 1 to 20 hectares) rather than accessing raster data directly. 

This modeling framework, coined VibrantForests, has been implemented with an initial set of modeling components to estimate fundamental forest structure attributes, with subsequent model stages to estimate forest product volumes and stocking levels relative to a site's carrying capacity. In this paper, we focus on the first stage of this modeling system: the estimation of several forest structure attributes that are fundamental to forest planning.

\section{Related Work}\label{sec:related}
An extensive body of research and scientific literature has emerged on the application of remote sensing for ``precision forestry'', but these systems remain fundamentally constrained by cost and logistical feasibility to relatively limited geographic extents and time windows. These include very-high-resolution sensor systems like laser scanning and photogrammetry collected by airplane, drone, and ground-based crews. These sensors provide precise and valuable data, but are infeasible for regular data updates and cannot support wall-to-wall applications beyond local-to-regional projects. We focus our discussion of related work on other efforts to produce regional-to-global forest mapping products at an annual or near-annual cadence. 

\subsection{Multi-Output Forest Modeling Systems}\label{sec:multi-output-models}
In contrast to more narrowly focused models for the estimation of carbon or aboveground biomass, there are far fewer examples of wall-to-wall forest modeling systems that aim to provide the more comprehensive set of forest attributes needed for operational forest planning and landscape management (e.g., basal area, merchantable timber volume, stocking density). We are not aware of any existing data products that produce this more comprehensive set of forest structure targets at an annual cadence or with coverage that extends beyond the United States, nor any that have adopted multi-task computer vision approaches. 

There are two primary examples of existing data products that provide more comprehensive forest structure metrics. The long-running LANDFIRE program \cite{rollins-2009, lapuma-2023} includes a large set of data layers tailored for wildland fuel and fire modeling and planning with periodic updates released every 2-5 years going back to 2011. LANDFIRE has evolved to include a variety of both imputation and direct regression and classification models, and to integrate numerous data sources including field data, lidar, and satellite imagery. These data have been primarily released in the form of wall-to-wall rasters at 30m resolution.

The other dataset offering more comprehensive coverage of forest attributes is TreeMap, which has provided wall-to-wall coverage of the contiguous United States (CONUS) since 2019 with predictions based on satellite imagery from 2014, 2016, 2020, and 2022 \cite{riley-2021}.  TreeMap data products are generated from a classification imputation model; it relies upon precise locations of US national forest inventory plots along with LANDSAT, climatic, and topographic features to predict the best-fitting inventory plot for each pixel \cite{riley-2021}. The TreeMap data product is distributed with several forest attributes released as wall-to-wall rasters at 30m resolution (currently offering more than 20 raster layers via the \href{https://lcms-viewer.fs2c.usda.gov/treemap}{TreeMap Explorer}). To allow analysts to derive arbitrary forest metrics from the underlying data, TreeMap is also provided as a plot identifier lookup layer. This lookup layer allows analysts to crosswalk additional detailed observations from field plots to wall-to-wall rasters, and is also what enables the use of TreeMap estimates as inputs for forward modeling inventory data using growth-and-yield models such as the Forest Vegetation Simulator (FVS, \cite{crookston-2005}). 

\subsection{Biomass-Focused Modeling Systems}\label{sec:carbon-models}
A handful of remote sensing data products have emerged that offer continental-to-global coverage of forest carbon or biomass at annual cadence and at moderate-to-high spatial resolution. These products have generally focused on the use of public satellite platforms including LANDSAT \cite{wulder-2022}, and more recently using GEDI \cite{dubayah-2020}, although the use of private satellite data sources is also notable in several recent efforts \cite{liu-2023, tolan-2024, anderson-2026}.

The extended history of data availability from the LANDSAT archive enabled products such as the eMapR aboveground biomass dataset, which spans the contiguous USA with annual updates from 1990-2018 \cite{kennedy-2018}. The spaceborne lidar platform GEDI, since its launch in 2018, has yielded partial spatial coverage with individual lidar footprints at \textasciitilde25m resolution, as well as wall-to-wall aboveground biomass data product at 1-kilometer resolution \cite{duncanson-2022} and has served as a source of training or validation in many newer computer vision models. 

\subsection{Models of Models}\label{sec:models-of-models}

The limited availability of confidently-located ground-truth data is a pervasive challenge for all large-scale modeling efforts to date. Modeling teams have taken different approaches to augment the limited number of field samples to allow new models to be trained for wall-to-wall inference. For example, the GEDI program addressed a common spatial mismatch problem: actual GEDI footprints rarely coincided with field inventory plots, so the team could not directly pair observed waveforms with ground-truth biomass. Instead, they simulated GEDI-like waveforms from existing airborne lidar surveys which did overlap with field plots, and then used these synthetic waveforms alongside allometric biomass estimates from field plots as training data \cite{duncanson-2022}. Within the USA, the field plots used in the GEDI training process were limited to NEON study sites and the application of allometric equations that only considered tree diameter and not height \cite{duncanson-2022}. 

The approach of generating an intermediate model that produces more training samples for a downstream model to learn from is now a common pattern. In particular, the apparent demand for high-resolution and frequently updated maps of forest biomass has attracted new modeling efforts from the private sector where global remote sensing data are used in combination with higher resolution aerial lidar to create training samples. Recent examples include \cite{liu-2023, tolan-2024, anderson-2026} where airborne lidar-derived rasters are employed as a training target for satellite-based models. 

The different modeling efforts to date vary in terms of whether they adopt the attributes sensed directly by lidar (e.g., canopy height) as a training target before applying allometric models, or whether they apply an allometric model to the lidar and use the results of that prediction as new training data (e.g., training on aboveground biomass estimates directly as targets). \citeauthor{liu-2023}~\cite{liu-2023} demonstrated the use of field observations collected by national forest inventory networks to generate allometric models that can then be applied to translate computer vision predictions of canopy height and cover to yield outputs of greater interest such as aboveground biomass (AGB). In contrast to an explicit allometric modeling stage, other projects like \cite{anderson-2026} utilize GEDI footprint-scale AGB estimates directly as training targets, essentially outsourcing the development of an allometric model by adopting the GEDI footprint-level data product as the training target for a computer vision model, which can allow the computer vision model to directly embed subtly different allometric relationships given spatial context.

\section{Study Area and Data}\label{sec:data}
\begin{figure}[htbp]
\centering
\includegraphics[width=\textwidth]{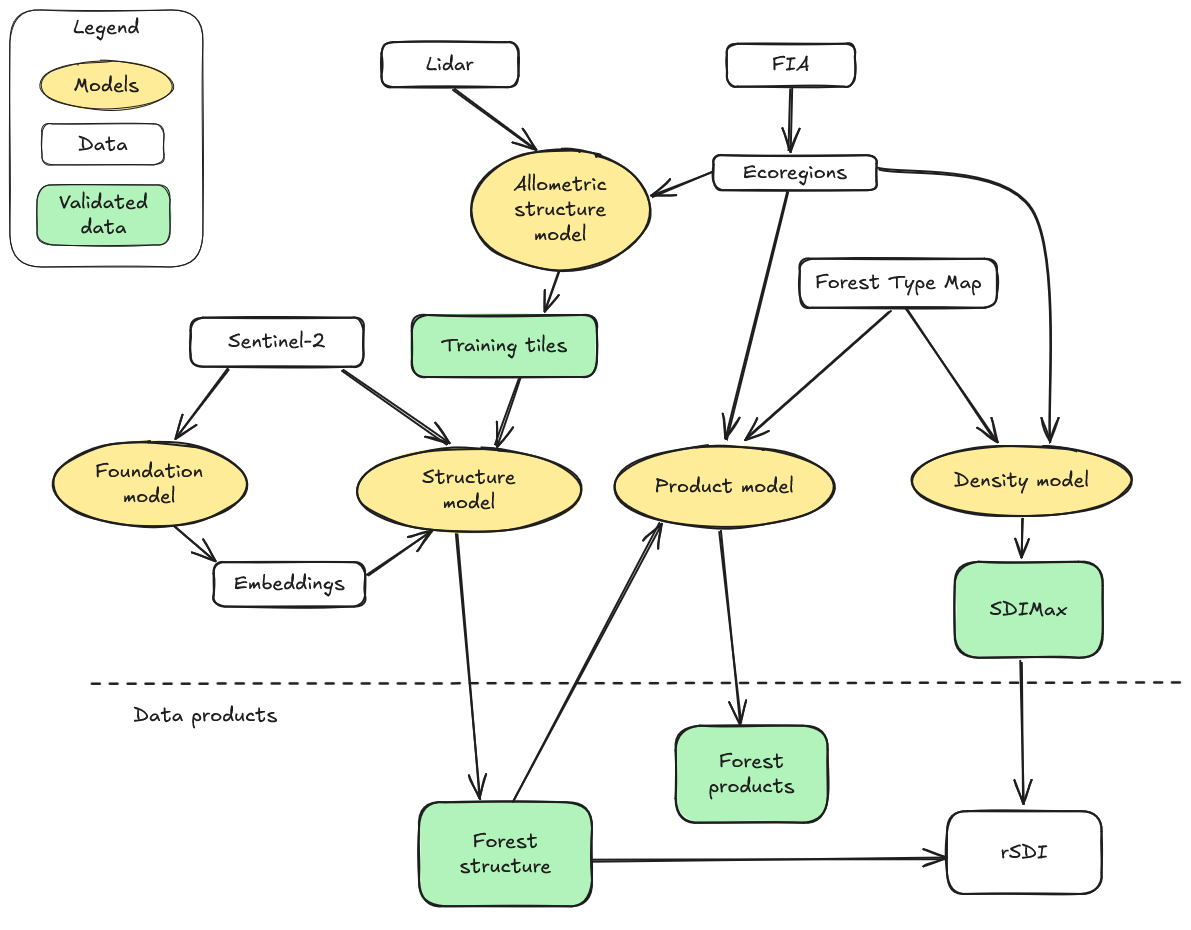}
\caption{Data flow through the VibrantForests framework. Inputs, including lidar, FIA, and Sentinel-2 data, feed two sequential modeling stages: an allometric structure model produces spatially explicit training tiles, and the satellite-based structure model that generates wall-to-wall forest structure predictions. This paper focuses solely on these two components (the left half of the diagram). Downstream forest product and density models are shown for context but are not discussed in detail in this article. More details about these components can be found in \href{https://github.com/Vibrant-Planet-Open-Science/Model-Cards/blob/main/model_cards/VibrantForests/ForestProducts.md}{ForestProducts} and \href{https://github.com/Vibrant-Planet-Open-Science/Model-Cards/blob/main/model_cards/VibrantForests/ForestDensity.md}{ForestDensity} Model Cards.}
\label{fig:vibrant-forests-diagram}
\end{figure}

We selected Sentinel-2 as our primary model input, as it provides cost-effective, consistent, wall-to-wall imagery with a high-frequency revisit cadence. To enable learning at fine spatial resolution, we generated high-fidelity training tiles over a representative subset of our target domain by training an allometric model on tabular summary statistics from national forest inventory plots. We then applied the allometric model to high resolution lidar-derived rasters to create training tiles. The flow of data and predictions through VibrantForests is illustrated in Figure \ref{fig:vibrant-forests-diagram}.

\begin{figure}[htbp]
\centering
\includegraphics[width=\textwidth]{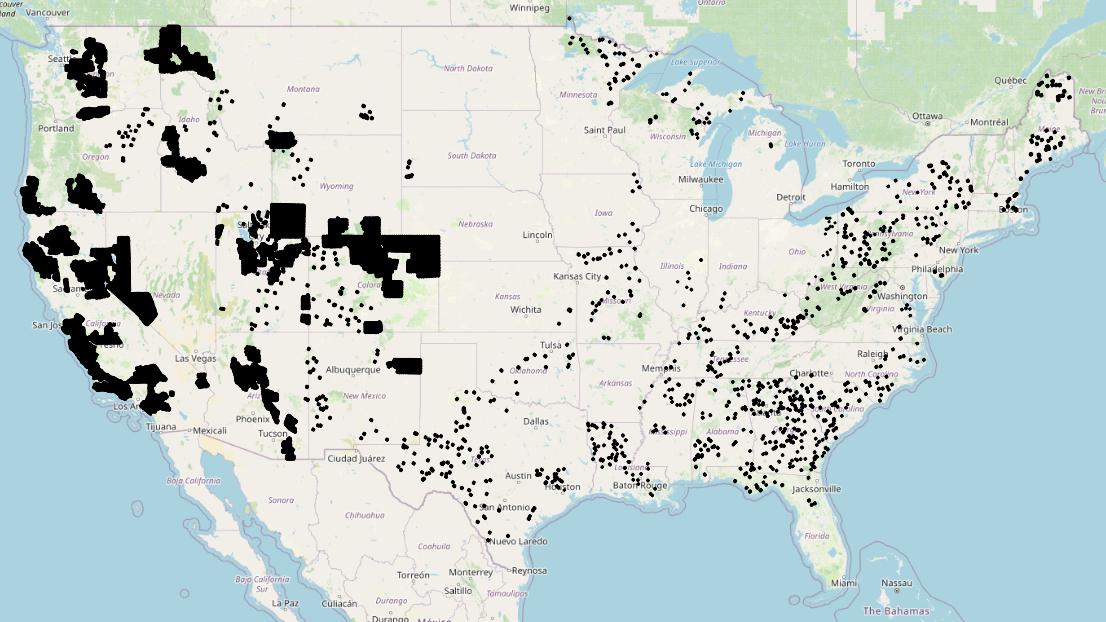}
\caption{The geographic extent of lidar-derived training data generated across CONUS. Lidar acquisitions concentrated in the western US used in model development by \cite{chang-2025} are complemented with a spatially balanced sample of new grid cells of lidar across CONUS, selected to match the density of forested national forest inventory plots in each ecoregion.}
\label{fig:lidar-footprint-across-study-area}
\end{figure}

We leverage previously-acquired lidar data employed in the development of a computer vision model for canopy cover and height prediction using aerial imagery \cite{chang-2025}. This prior modeling effort had a focus on Wildfire Crisis Strategy landscapes, defined by the United States Forest Service to prioritize regions at the greatest risk of severe wildfire and produced a dense sample of lidar from the western US \cite{-n} (see Figure \ref{fig:lidar-footprint-across-study-area}).

\subsection{Grid System and Extent}\label{sec:grid-system}

\begin{figure}[htbp]
\centering
\includegraphics[width=\textwidth]{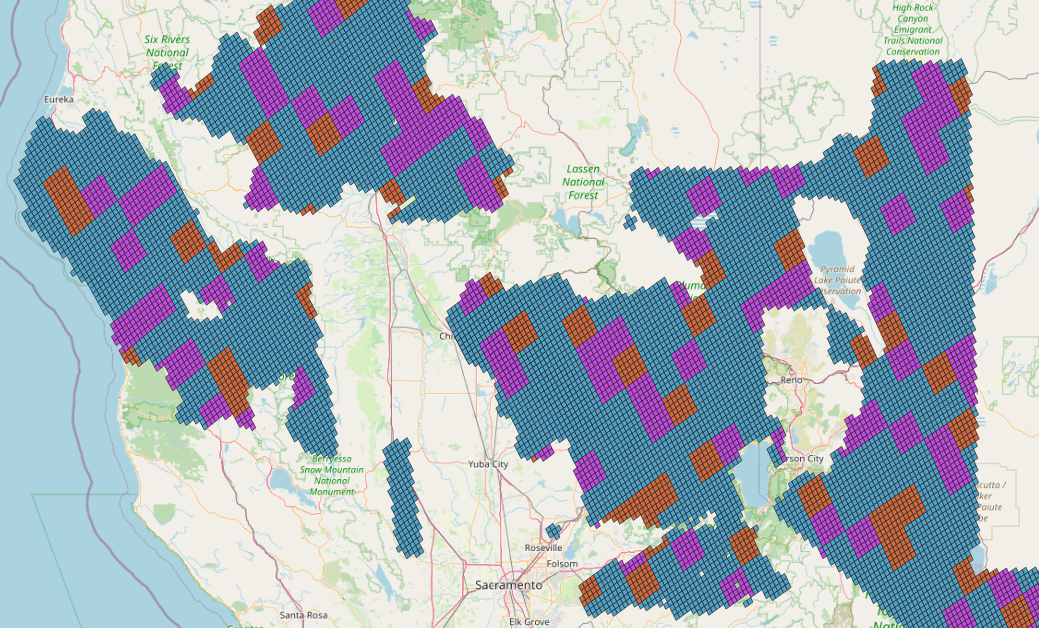}
\caption{Spatial partitioning of 2,500m × 2,500m tiles into training (blue), validation (purple), and test (orange) sets based on 15km x 15km parent grid cells. Partitioning at the parent-cell scale mitigates data leakage between sets due to spatial autocorrelation.}
\label{fig:tile-partitioning-scheme}
\end{figure}

All incoming raster and point cloud data were reprojected to EASE Grid 2.0 North (EPSG:6931) and tiled into 2,500m × 2,500m samples (with 30m padding on each side, yielding 2,560m × 2,560m input tiles). To prevent spatial data leakage between partitions during model training and evaluation, tiles were pre-assigned to partitions based on their parent 15km × 15km grid cell (see Figure \ref{fig:tile-partitioning-scheme}).

\subsection{National Forest Inventory}\label{sec:fia}

The US Forest Service's Forest Inventory and Analysis (FIA) program provides a systematic and spatially balanced inventory of forests across CONUS and US islands while a sparser sampling strategy is applied in Alaska \cite{gray-2012}. The geographic expanse and diversity, sample size, and revisit frequency of the FIA network provides a robust data source for model development and validation. However, precise FIA plot locations are not available. We therefore restricted our use of FIA data to publicly-available attributes and designed our lidar-based models to operate on non-spatial tabular data only.

\begin{SCfigure}[1.0][htbp]
\includegraphics[width=0.5\textwidth]{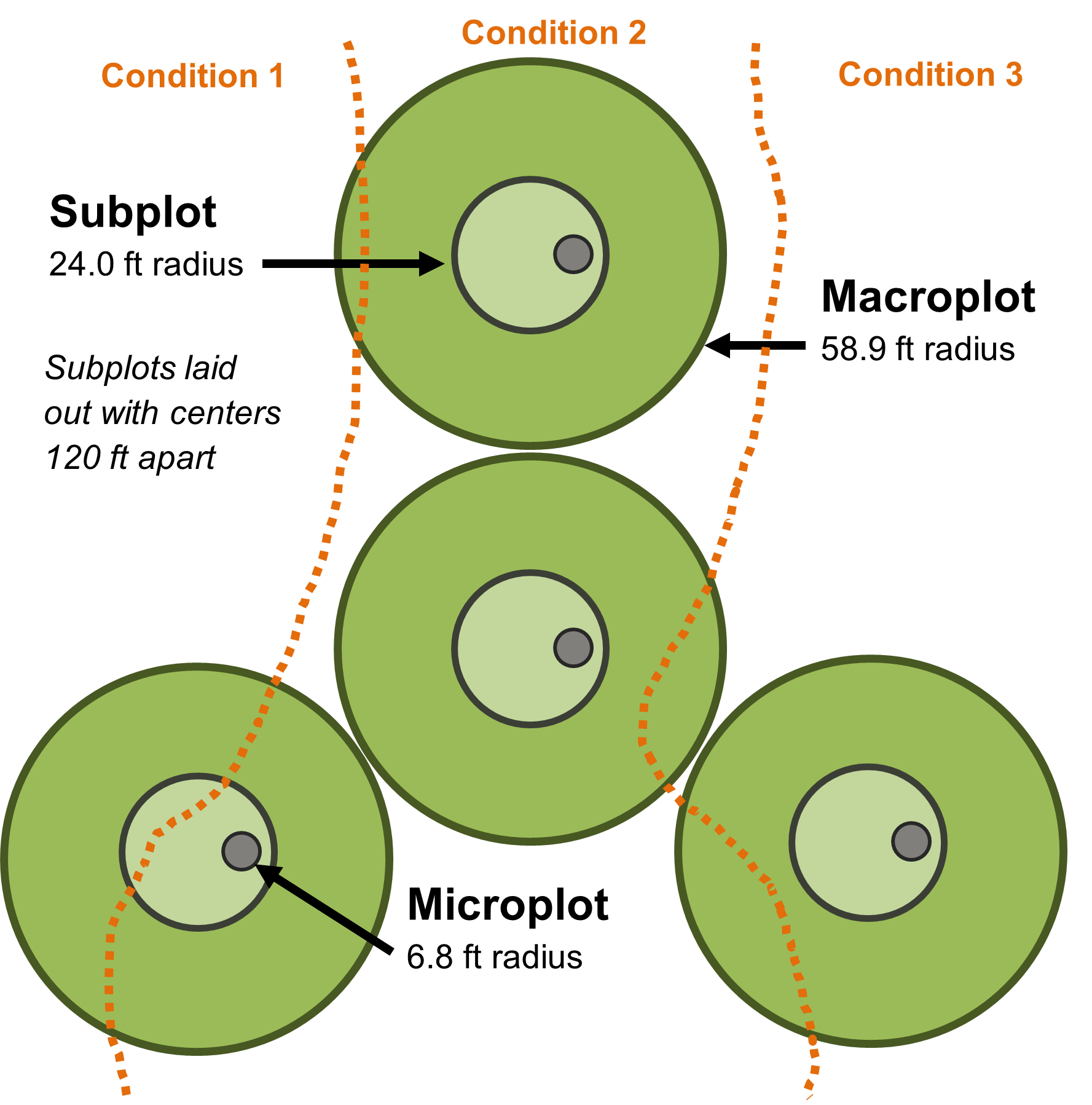}
\caption{Illustration of three distinct ``conditions'' delineated across the four-subplot layout of an FIA plot. Adapted from FIA documents by \cite{diaz2024}. The footprint of the four subplots and associated macroplots combine to be considered the footprint of an FIA plot. The subplot-specific footprint is relied upon in this study as the spatial scale for inventory data extraction and creation of allometric models.}
\label{fig:fia-plot-layout}
\end{SCfigure}

The FIA subplot was chosen as a key sampling concept in our development of training data. The contemporary FIA plot design, which was adopted consistently nationwide beginning around the year 2000, is a cluster of four circular subplots, each with a radius of \textasciitilde7.3m (24ft; Figure \ref{fig:fia-plot-layout}). The footprint of an individual subplot covers an area of \textasciitilde168m², which is moderately larger than the size of the 100m² pixels generated by Sentinel-2. Nevertheless, we considered subplot-level scale sufficiently similar while still providing a meaningful scale at which forest structure attributes can be reasonably summarized. By choosing to use subplots as the spatial scale for summarizing inventory data, it was possible for us to map target variables at a much finer scale and with more meaningful texture (i.e., high-resolution variation in target forest attributes) than summaries at the plot-level would have supported.

Training tiles for the satellite forest structure model were created by applying an FIA-derived allometric model to high-resolution lidar-derived rasters of canopy cover and canopy height. 

\subsection{USGS 3DEP Lidar}\label{sec:3dep}

Publicly-available lidar point clouds were sampled from the USGS 3DEP program \cite{usgs-2026}, limited to acquisitions with a Quality Level rating of 2 or better. We processed the point clouds using the Point Data Abstraction Library (PDAL) \cite{butler-2021, butler-2024} to remove points classified as noise by the vendor and any points that were more than 2m below or 90m above vendor-classified ground surface. Point clouds were processed into three raster layers: a Canopy Height Model (CHM) representing maximum vegetation height at 0.5m resolution, canopy cover layers derived from height-stratified return counts at 10m resolution, and a Digital Terrain Model representing ground elevation. Canopy cover above 2m was calculated as the ratio of returns above 2m to total returns within each pixel at 10m resolution.

\subsection{Sentinel-2 Imagery}\label{sec:sentinel-2}

The satellite forest structure model uses 12-band Sentinel-2 Level-2A surface reflectance imagery \cite{drusch-2012}, specifically the Sentinel-2-C1-L2A collection distributed by Element84 \cite{-2026a}. We aggregated observations via a temporal pixel-wise median of cloud-masked images collected within June-July-August of a given year. Individual images were masked using the Scene Classification Layer (SCL) to exclude pixels classified as no data, defective, cloud shadow, cloud probability medium, cloud probability high, or thin cirrus. This temporal median aggregation reduces transient artifacts (e.g., residual cloud contamination) and captures vegetation structure during peak growing season. Inputs from all Sentinel-2 bands were processed and resampled (nearest neighbor) to 10m resolution in the EASE grid 2.0 projection. 

\subsection{Pacific Northwest Field Plots}\label{sec:remote-sensing-plots}

\begin{SCfigure}[1.0][htbp]
\includegraphics[width=0.5\textwidth]{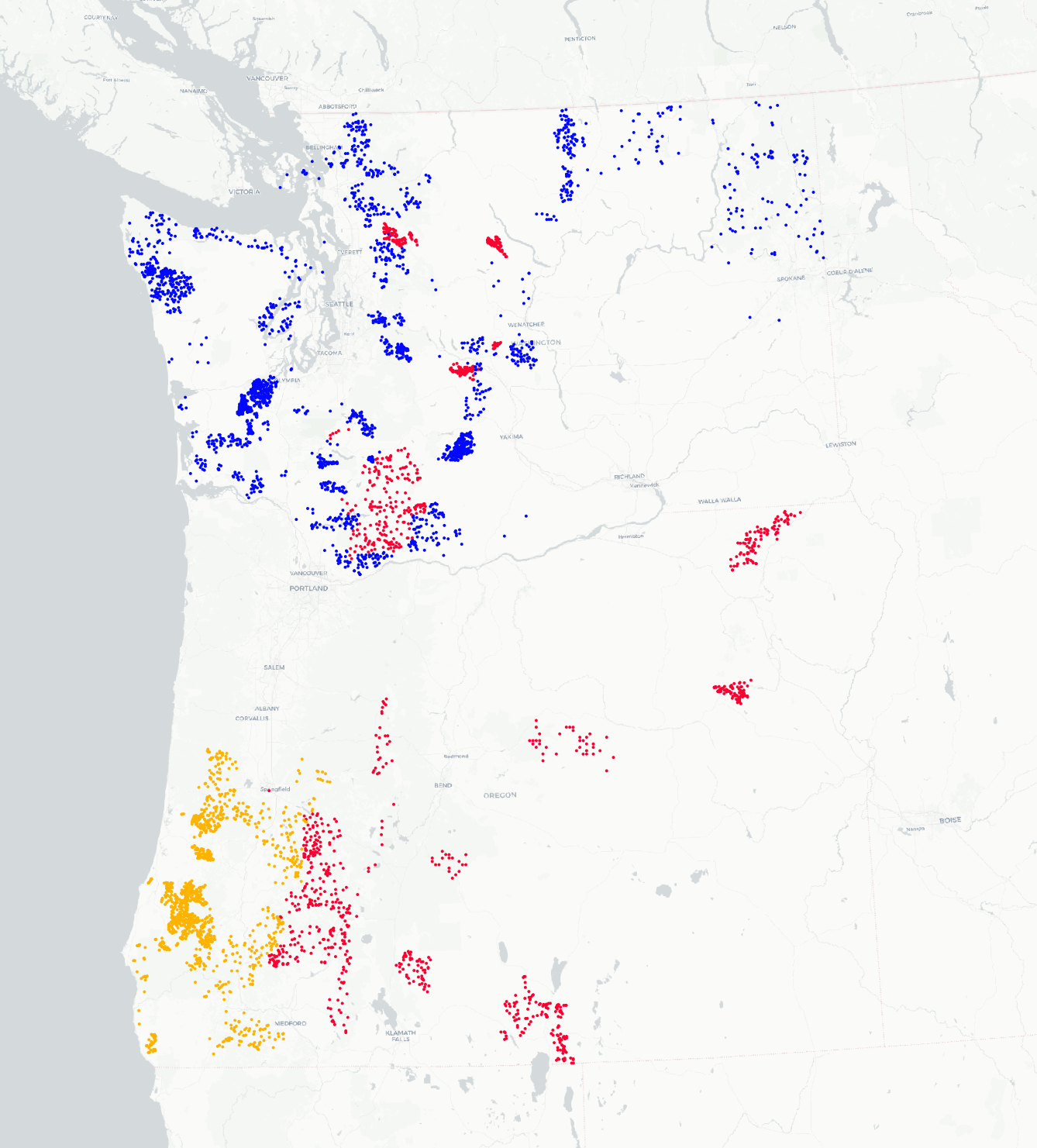}
\caption{The distribution of Pacific Northwest field plots used for validation. Points in blue were collected by Washington DNR (n=2,487), points in red by US Forest Service (n=1,223), and points in orange by BLM (n=1,379).}
\label{fig:pnw-remote-sensing-plots}
\end{SCfigure}

Forest plot data collected by the Bureau of Land Management (BLM; Oregon), US Forest Service (USFS; Oregon \& Washington), and Washington Department of Natural Resources (DNR) from 2010 through 2018 were compiled through a combination of direct correspondence and public records requests (see Figure \ref{fig:pnw-remote-sensing-plots}). The data include raw field measurements with plot locations, as well as tree species, diameter, height, and live/dead status among several other attributes. From the 5,089 field plots, 551 were removed from validation analyses due to the absence of co-registered lidar available at the time of data acquisition.

Plot sizes and sampling protocols varied among agencies, as did the dates of field data collection. In general, BLM and USFS conducted regional field campaigns to correspond with the collection of aerial lidar over specific National Forests or BLM Districts. Washington DNR collected field plot data to correspond with statewide NAIP imagery collected in 2021 that was processed using photogrammetry to generate point clouds.

These data are rare examples of confidently-located plot-level forest observations in managed forest landscapes that can be used as ground-truth. We have identified several other sources of plot- and stand-based measurements in managed forests, reserves, and research sites and will continue to collect and ingest additional field measurements over time to improve the diversity and depth of datasets available for model evaluation.

To avoid introducing further error and bias, field plots were not grown forward with the Forest Vegetation Simulator (FVS) as had been adopted in an earlier modeling effort using these plots \cite{diaz2024}. Re-analysis of earlier growth-and-yield simulations for these plots by \cite{diaz2024} revealed dozens of plots near the maximum Stand Density Index boundary used by FVS where the model triggered significant mortality as it grew the plot data forward which could not be empirically validated. We thus limited the use of FVS to calculate BA, QMD, and AGB at the time of field measurement and accepted the time lag between field observations and the acquisition of imagery used in model predictions as a more consistent and interpretable source of error than growth-and-yield modeling would have added.

\section{Methods}\label{sec:method}

\subsection{Allometric Forest Structure Model}\label{sec:allometric-structure-model}
The allometric forest structure model is a multi-target tabular regression pipeline that predicts forest structure attributes from vegetation structure summaries. It was not intended as a standalone inference model, but rather to generate spatially explicit training labels for the satellite forest structure model (described in \nameref{sec:forest-structure-model}). The allometric pipeline learns relationships between FIA subplot-level measurements and canopy/terrain metrics, then applies those learned relationships to lidar-derived rasters to produce training tiles at 10m resolution.

The allometric forest structure model predicts three target attributes:
\begin{itemize}
    \item AGB: aboveground live tree biomass [Mg/ha]
    \item BA: basal area of live trees [m²/ha]
    \item QMD: quadratic mean diameter [cm]
    
\end{itemize}

Four features are inputs to those predictions:
\begin{itemize}
    \item Cover: canopy cover [\%]
    \item Height: maximum canopy height [m]
    \item Elevation: elevation above sea level as reported for the central subplot [m]
    \item Ecoregion: identifier of the EPA Level IV ecoregion containing the sample
\end{itemize}

\subsubsection{Architecture}\label{sec:allometric-architecture}
The allometric forest structure model was implemented as a scikit-learn \cite{pedregosa2011} pipeline with a \texttt{MultiOutputRegressor} of \texttt{GradientBoostingRegressor} instances for predicting each target attribute.

\subsubsection{Data Processing}\label{sec:allometric-datasets}
Subplot-scale training data were prepared from the FIA database and partitioned into training (n=148,262), validation (n=43,470), and testing (n=21,361) sets.

We extracted subplot-scale summaries from the FIA database for all plots where live trees were observed. Selected attributes included canopy cover (CC), maximum tree height, basal area (BA), aboveground biomass (AGB), and quadratic mean diameter (QMD). Field data for all trees with diameter at breast height $\ge$ 1 inch observed within FIA subplot footprints were incorporated to fit an allometric model that predicted AGB, QMD, and BA as a function of field-measured canopy cover, canopy height, elevation, and ecoregion. Canopy cover is recorded for ``conditions'' which capture transitions between forest or land cover types across the four-subplot cluster layout in the FIA sampling design (see Figure \ref{fig:fia-plot-layout}). At the subplot-scale, canopy cover was calculated as a weighted average based on the proportion of the subplot covered by each condition. We calculated canopy height as the maximum height of trees observed within the subplot footprint. For elevation, we used the point estimate reported in the FIA database. The EPA Level IV ecoregion for each sample was defined using a spatial join of the fuzzed FIA plot coordinates with the EPA Level IV ecoregions layer \cite{usepa-2015}. Although fuzzed locations may result in some plots being assigned to a neighboring ecoregion, allometric relationships are unlikely to shift dramatically across these boundaries, so we expect any resulting noise to be minimal.

We only kept FIA samples where all live trees included in the subplot-level summary metrics could be confidently restricted to fall within the subplot footprint (and used the subplot area as the factor for expanding tree observations to an areal basis). Subplots that contained no live trees we discarded. In some regions, the FIA program employs macroplots to sample large live trees and snags. In regions where macroplots are used, subplots were discarded in cases where the distance to any tree could not be determined if the tree's diameter was large enough such that it could have been observed outside the subplot (but within the macroplot). We relied upon archived releases of FIA databases, which included bearings and distances to measured trees to mitigate the censorship of tree locations being applied in contemporary versions of the FIA database. A total of 213,093 subplot-level samples with valid data were produced. 

The allometric forest structure model is implemented with a preprocessing stage including a standard scaler for numerical features (which transforms features to have a mean of zero and unit variance) and one-hot encoding for categorical features. 

\subsubsection{Training}\label{sec:allometric-training}
The allometric forest structure model was written to allow alternative multi-output architectures to be evaluated during model fitting, including independent univariate models, a \texttt{MultiOutputRegressor}, and a \texttt{RegressorChain}. The individual models we considered were \texttt{RandomForestRegressor} and \texttt{GradientBoostingRegressor}. 

Model training and hyperparameter tuning was executed using grid search cross-validation to adjust base-model settings such as number of trees and depth of trees, learning rate, minimum number of samples per split and leaf node, etc. Mean squared error was used as the loss function. During cross-fold validation, the model pipeline's performance for each hyperparameter configuration was evaluated on the validation split with per-target metrics calculated including mean absolute error, root mean squared error, and the coefficient of determination ($R^2$). A \texttt{MultiOutputRegressor} of \texttt{GradientBoostingRegressors} was identified as the optimal model architecture, with hyperparameter settings of learning rate=0.05, max depth=7, max features=``sqrt'', minimum samples per leaf=4, minimum samples per split=10, number of trees=300, and subsample=0.8.

\subsubsection{Inference}\label{sec:allometric-inference}
The allometric forest structure model was applied to lidar-derived rasters to generate training tiles for the satellite forest structure model. In this context, lidar-derived canopy-cover, canopy height, and elevation are treated as direct proxies for canopy cover, height, and elevation derived from FIA subplots. To compute subplot-scale equivalent metrics from the lidar data, we passed a circular kernel the same size as an FIA subplot over the CHM and canopy cover rasters. Within each kernel, we retrieved the maximum value of canopy height and the mean value of canopy cover. 

\begin{figure}[htbp]
\centering
\includegraphics[scale=0.99]{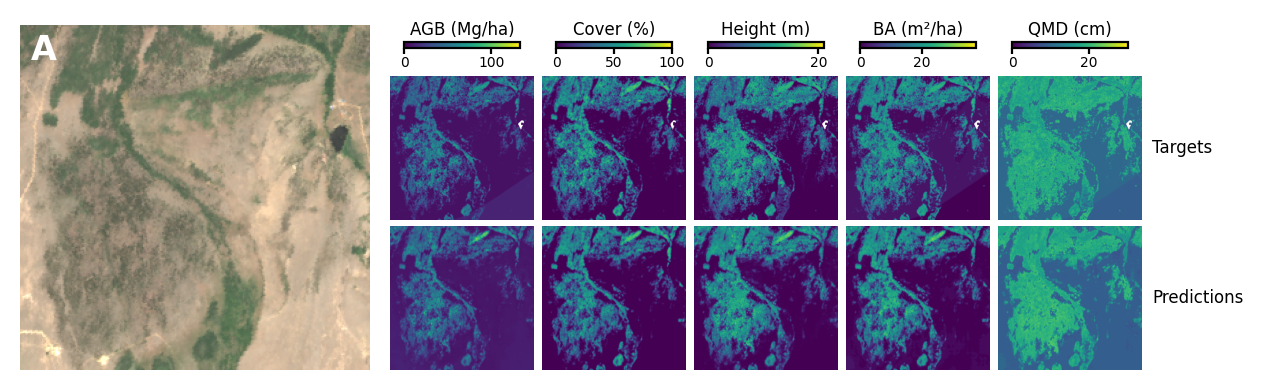}
\includegraphics[scale=0.99]{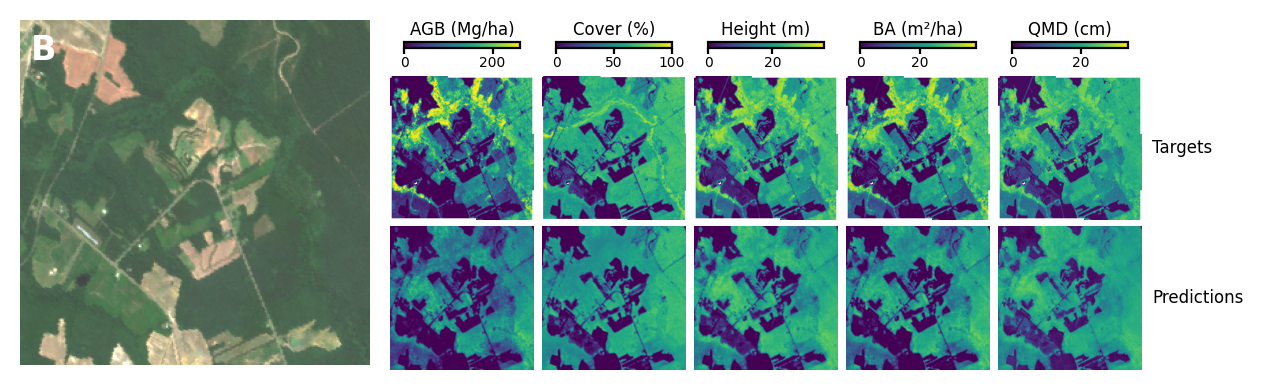}
\includegraphics[scale=0.99]{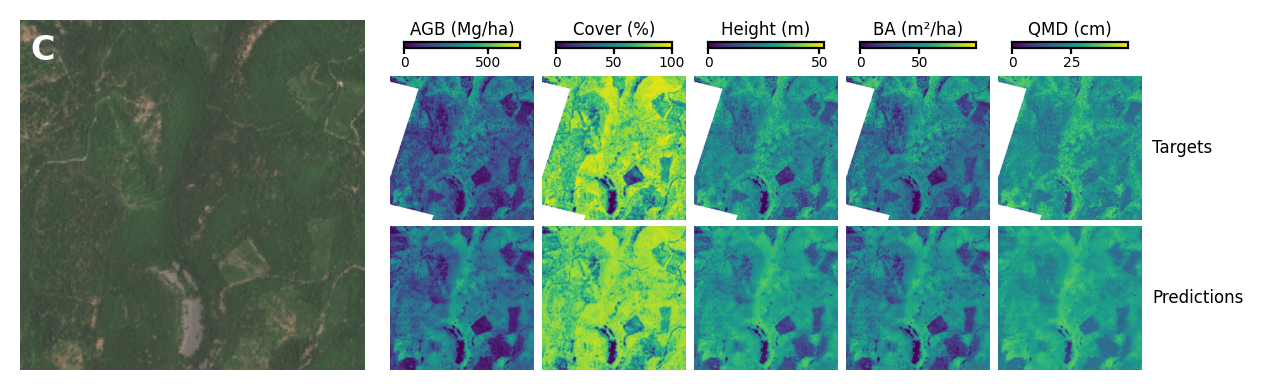}
\includegraphics[scale=0.99]{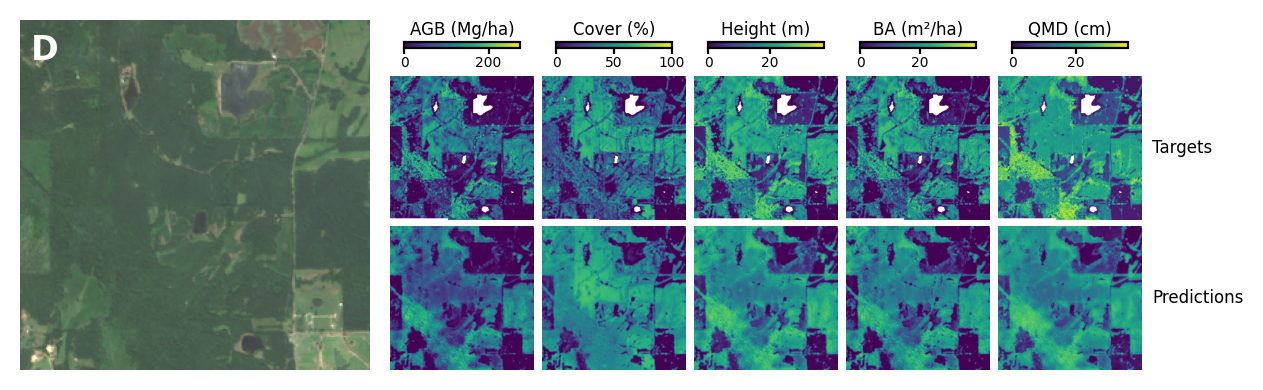}
\caption{Lidar-derived targets and satellite-based predictions for several 2560m$\times$2560m examples. Sentinel-2 natural color image from June-August composite at left. Tiles to right of each image show targets (top) and predictions (bottom). Color bars for each target/prediction pair are shared, but ranges differ among example locations. Examples are from southern Wyoming (A), eastern North Carolina (B), south-central Washington (C), and central Mississippi (D). All examples are from the test partition.}
\label{fig:training-and-prediction-pairs}
\end{figure}

The allometric forest structure model was applied to the lidar-derived layers, along with rasterized versions of global ecoregion and biome identifiers, to estimate the modeled target forest structure attributes for each pixel at 0.5m resolution. Lidar-derived canopy height and canopy cover were then combined with high-resolution predictions of basal area, quadratic mean diameter, and aboveground biomass by the allometric forest structure model before being resampled to 10m resolution to be used as target layers for training the satellite forest structure model (see Figure \ref{fig:training-and-prediction-pairs}).

In contrast to \cite{liu-2023}, we apply the allometric forest structure prior to training the computer vision model. The allometric model is applied to generate predictions of additional target variables across tiles, which are then used in combination with lidar-derived canopy height and cover as training targets for the multi-target satellite-based forest structure model. Rather than applying the allometric model as a post-processing step (i.e., to noisy per-pixel predictions), we used it to generate spatially continuous training targets upfront. Our intent was to allow the satellite-based computer vision model to implicitly learn allometric relationships between target variables with spatial context. This approach encourages the computer vision model to learn relationships between canopy cover, canopy height, and the target variables (AGB, BA, and QMD) that are spatially coherent, potentially differing subtly from the non-spatial relationships learned by the allometric model, which was trained only on tabular data. 

\subsection{Satellite Forest Structure Model}\label{sec:forest-structure-model}

The satellite forest structure model is a multi-target regression model used to directly estimate five fundamental forest structure attributes at 10m resolution from satellite imagery:
\begin{itemize}
    \item AGB: aboveground live tree biomass [Mg/ha]
    \item Cover: canopy cover [\%]
    \item Height: canopy height [m]
    \item BA: basal area of live trees [m²/ha]
    \item QMD: quadratic mean diameter [cm]
\end{itemize}

Trees per hectare (TPH) and Stand Density Index (SDI) are then derived deterministically from predicted BA and QMD:

\begin{equation}
    \text{TPH} = \frac{\text{BA}}{\text{QMD}^2 \times 0.00007854}
\end{equation}
\begin{equation}
    \text{SDI} = \text{TPH} \times \left( \frac{\text{QMD}}{25.4} \right)^{1.605}
\end{equation}

Where:
\begin{itemize}
    \item[] $\text{TPH}$ = live trees per hectare
    \item[] $\text{BA}$ = subplot-level basal area in live trees ($\text{m}^2/\text{ha}$)
    \item[] $\text{QMD}$ = quadratic mean diameter ($\text{cm}$)
    \item[] $0.00007854$ = the metric forester's constant ($\frac{\pi}{40,000}$)
    \item[] $1.605$ = Reineke's power defining the slope of the maximum size-density relationship
    \item[] $25.4$ = standard reference diameter ($\text{cm}$)
\end{itemize}

\subsubsection{Architecture}\label{sec:structure-architecture}

The satellite forest structure model uses a Feature Pyramid Network (FPN, \cite{lin2017}) architecture that follows from a Vision Transformer-based Masked AutoEncoder \cite{he2022}. In brief, the Masked AutoEncoder is an image transformer trained to predict missing sub-tiles of Sentinel-2 images. The bottleneck of the Masked AutoEncoder holds a compressed representation of Sentinel-2 imagery that embeds spatial and spectral context information as a vector of features per image patch. After training, the Masked AutoEncoder's encoder is utilized as a frozen feature extractor, producing multi-scale feature representations (embeddings) at multiple transformer depths. These embeddings are then supplied alongside Sentinel-2 imagery as features through the FPN with a pixel-wise regression head to predict the five forest structure attributes. The satellite forest structure model was trained on 256$\times$256 pixel (2560$\times$2560 meters) images, enabling the model to capture spatial patterns from local- to stand-scales. 

\subsubsection{Data Sources \& Processing}\label{sec:structure-datasets}

\paragraph{Features}
The satellite forest structure model uses a seasonal mosaic of Sentinel-2 satellite imagery as its sole input. The seasonal mosaic for a given year was passed through the frozen encoder of the Masked AutoEncoder to generate embeddings at 80m resolution. Internally, the satellite forest structure model combines these embeddings with the original Sentinel-2 composite to predict forest structure attributes. 

\paragraph{Targets}
The satellite forest structure model was trained using tiles at 10m resolution containing a combination of lidar-derived canopy height and cover along with the allometric-predicted AGB, BA, and QMD layers described in section \ref{sec:allometric-inference}. Target variables were normalized to a [0, 1] range using min-max normalization with target-specific minimum and maximum values to ensure that all five forest structure attributes—which have different scales and ranges (e.g., AGB [0, 2936] Mg/ha range versus cover [0, 100] \% range)—contribute equally to the loss during training.

\subsubsection{Training}\label{sec:structure-training}

We trained the satellite forest structure model using PyTorch Lightning \cite{falcon-2020} for up to 150 epochs with early stopping based on monitored validation loss (patience of 10 epochs). The model was trained as a quantile regression model, outputting predictions for five quantile levels (0.1, 0.3, 0.5, 0.7, 0.9) for each target. This approach allows the model to explicitly learn multiple points reflected the shape of the target distribution (10th, 30th, 50th, 70th, and 90th percentiles), providing built-in uncertainty quantification. Unlike standard regression which outputs only a single point estimate (typically the mean), quantile regression can be applied to capture the shape and spread of the distribution, which is particularly useful for the right-skewed distributions of our target variables.  We computed per-target losses as the average loss across all quantile levels. The final loss was computed as the average of per target losses across all five forest structure attributes.

We used the AdamW optimizer \cite{loshchilov2018} with a learning rate of 1e-4 with a step decay scheduler (step size of 10 epochs and gamma of 0.7), weight decay of 0.05, and betas of (0.9, 0.999). The loss was scaled based on the ratio of valid pixels to account for spatial variation in data availability. During training, we evaluated validation loss every three epochs against the validation partition. 

\subsubsection{Inference}\label{sec:structure-inference}

Satellite forest structure outputs were generated as 256$\times$256 pixel rasters using a sliding overlapping window approach. To eliminate edge effects, the final 15km mosaics were computed as a weighted average of overlapping tiles using a Gaussian kernel centered on each tile. This approach up-weights pixels near the tile center, where model predictions are more confident, and down-weights pixels near edges. In our reporting of model performance below, we rely upon the median (50th percentile) prediction of each forest structure attribute in each pixel.

\subsection{Model Evaluations}\label{sec:model-evaluations}

Model performance is reported using Mean Absolute Error (MAE), Root Mean Squared Error (RMSE), mean bias, the coefficient of determination ($R^2$), and Pearson's $r$. Qualitative assessments were conducted through visual inspections of scatter plots of predicted versus observed values.

We chose to evaluate the wall-to-wall predictions from the satellite forest structure model against field data based on satellite imagery collected in 2024. This vintage of imagery was chosen to leverage wall-to-wall data layers that were generated across CONUS representing 2024 forest conditions for inclusion in the \href{https://forestinnovationplatform.org/}{Forest Innovation Platform} being developed by American Forests. 

\subsubsection{Pacific Northwest Field Plots}\label{sec:plot-evaluation}

Field plot measurements were compiled using the appropriate regional variant of FVS to calculate a broad suite of attributes from the raw inventory data. We did not utilize canopy cover estimates generated by FVS, which have been shown to be systematically biased compared to field-based and remote sensing methods \cite{gray-2021}. Instead, we utilized lidar-derived estimates of canopy cover for each plot footprint from the most recent lidar collection available—if any—over each plot.

As noted previously, field plots were not grown forward due to potential bias introduction, and the use of FVS was thus limited to calculate plot attributes at the time of measurement with acceptance of the time lag between field observations and the acquisition of imagery used in model predictions.

\subsubsection{Regional Distributions}\label{sec:regional-evaluation}
For evaluation data, we extracted forest structure attributes from FIA plots including both forested and non-forested plots. We chose to utilize plot-level summaries for evaluation purposes given the intended uses of model predictions at the coarser scale of management units (as opposed to focusing evaluation on noisier summaries of forest structure at the subplot scale).

Forest structure attributes were set to zero where FIA plots were recorded as non-sampled due to absence of forest cover, acknowledging this omits cover by trees outside of forests and by non-tree vegetation. We aggregated wall-to-wall predictions from satellite-based model on 2024 imagery to hexagon-level means, and made comparisons with FIA observations on an all-lands basis (without filtering by land cover type).

To facilitate comparison with other data providers, we included an evaluation of our forest structure predictions against the \citeauthor{menlove-2020}~\cite{menlove-2020} AGB data layer where FIA observations from 2009-2019 were averaged to the scale of 64,000-hectare hexagons. Because \citeauthor{menlove-2020} \cite{menlove-2020} only summarized AGB, we also queried the FIA database for the most recent 10 years of data available (2012-2022) to evaluate hexagon-level performance for all our target variables. Hexagon assignments for each plot were based on the hexagon IDs linked to each plot in the FIA database. This query includes more recent FIA samples and additional hexagons that were not represented in the \citeauthor{menlove-2020} \cite{menlove-2020} dataset.

\section{Results}\label{sec:results}

\subsection{Evaluation on Test Partition}\label{sec:train-test-results}

\begin{figure}[htbp]
\centering
\includegraphics[scale=0.6]{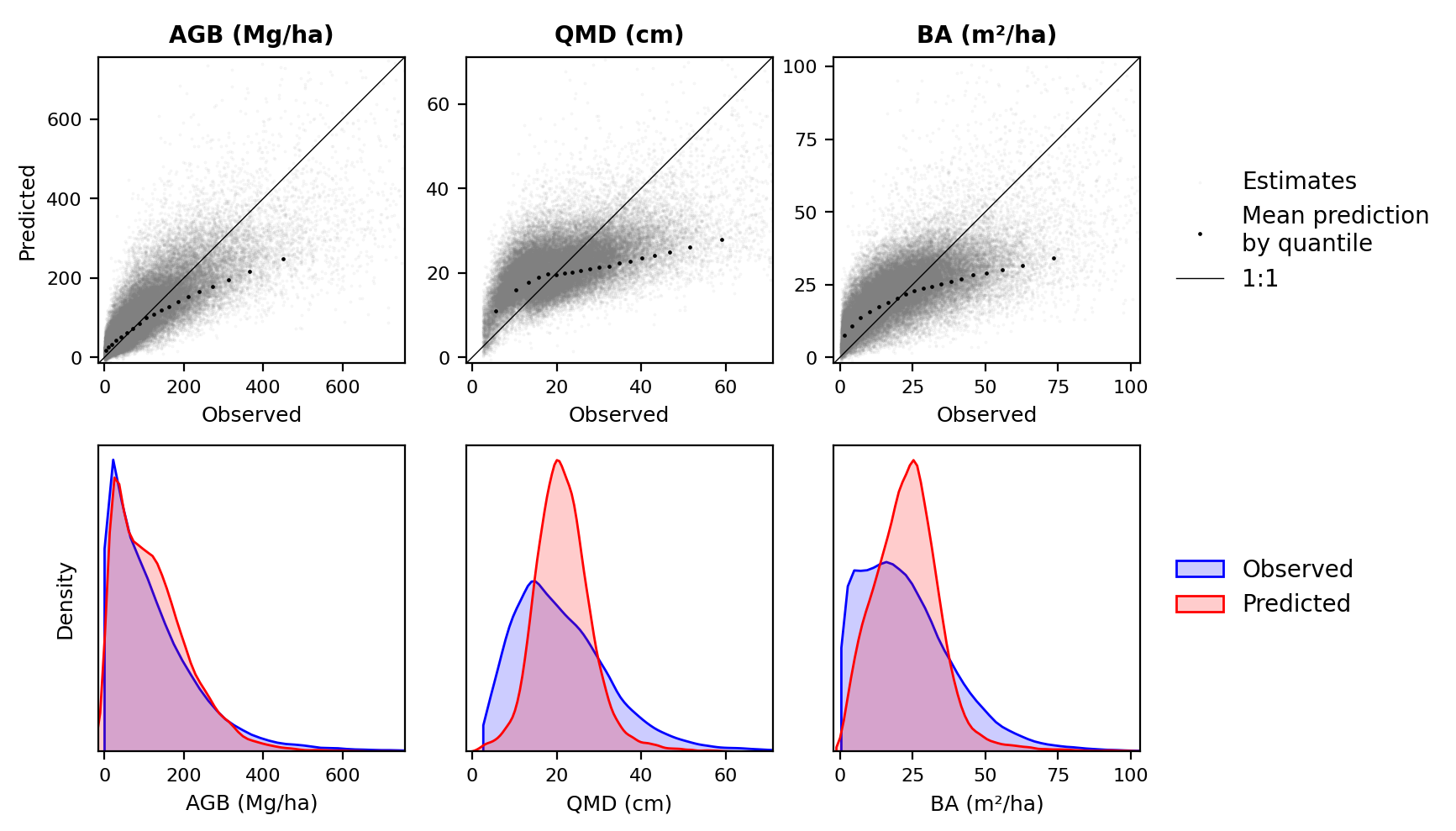}
\caption{Allometric model performance on the FIA subplot test partition (n=43,470). Scatter plots (top) show predicted vs. observed values with the 1:1 line. The black points in each scatter plot show the average value predicted for each quantile of observations, spanning the 5th through 95th percentiles in steps of 5. Density plots (bottom) compare distributions of observed and predicted values.}
\label{fig:allometric-structure-subplot-level-scatter}
\end{figure}

\paragraph{Allometric Forest Structure.}
The allometric forest structure model provided reasonable predictions of AGB, QMD, and BA on FIA plots held out from training and hyperparameter tuning (see Figure \ref{fig:allometric-structure-subplot-level-scatter} and Table \ref{tab:allometric-structure-test-table}). The allometric model began to lose the AGB signal in dense and tall canopies, leading to under-prediction biases above 150-200 Mg/ha. Due to the rarity of AGB observations above this range, the distribution of predicted AGB values was largely consistent with observed AGB values far into the right tail of the distribution. 

For both BA and QMD, the allometric model tended to regress more strongly to the mean, with predictions covering a compressed distribution compared to observations. BA and QMD predictions generally showed low bias, but they tended to be flatter across the range of observed values, explained less of the variance among observations compared to AGB, and demonstrated over-prediction bias in the low range of values and under-prediction bias in the high range. 

\begin{table}[htbp]
\centering
\caption{Summary statistics from evaluation of the allometric model against FIA subplots in the test partition. Obs and Pred columns display the mean ± standard deviation for observations and predictions, respectively.}
\label{tab:allometric-structure-test-table}
\begin{tabular}{lccccccc}
\toprule
Target & MAE & RMSE & Mean Bias & $R^2$ & Pearson's $r$ & Obs & Pred \\
\midrule
AGB (Mg/ha) & 47.8 & 79.9 & 14.2 & 0.66 & 0.82 & 125 ± 138 & 126 ± 113 \\
BA (m²/ha) & 9.5 & 13.4 & 1.9 & 0.49 & 0.70 & 24 ± 19 & 24 ± 13 \\
QMD (cm) & 7.3 & 9.8 & -2.3 & 0.38 & 0.62 & 22 ± 12 & 22 ± 7 \\
\bottomrule
\end{tabular}
\end{table}

\begin{figure}[htbp]
\centering
\includegraphics[width=\textwidth]{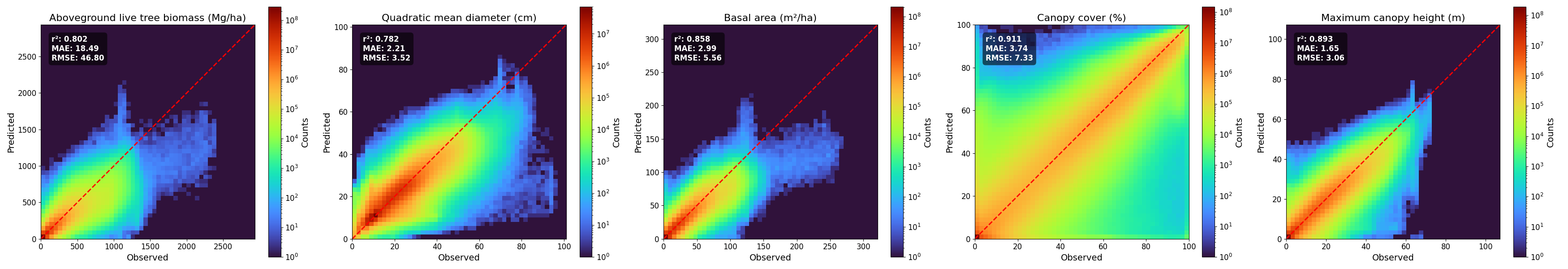}
\caption{Satellite model performance on lidar-derived test tiles, shown as two-dimensional histograms with pixel values representing log-scaled counts where red is high density and blue is low density. The dashed red diagonal is the 1:1 line. The dense cluster near the origin in most panels reflects an abundance of non-forested pixels (zero AGB, BA, etc.). Performance metrics are inset in each panel. High $R^2$ reflects ability of the satellite model to explain variation in allometric model predictions, not direct accuracy against field plots (see \nameref{sec:plot-scale-results}).}
\label{fig:forest-structure-pixel-level-scatter}
\end{figure}

\paragraph{Satellite Forest Structure.}\label{sec:structure-train-test}
We evaluated satellite model predictions against tiles in the test partition produced by the allometric model  (see Figure \ref{fig:forest-structure-pixel-level-scatter}). For every target variable, the model explained the vast majority of variation in the test tiles with very limited bias, with QMD showing the lowest $R^2$ value at 0.78.

For forest attributes detected directly by lidar, we observed strong performance at recovering lidar-derived canopy attributes:
\begin{itemize}
    \item Cover: MAE = 3.7\%, $R^2 = 0.91$
    \item  Height: MAE = 1.65m, $R^2 = 0.89$
\end{itemize}

The differences between $R^2$ values among pixel-scale evaluations (see Figure \ref{fig:forest-structure-pixel-level-scatter}) versus plot-scale evaluations (see Figure \ref{fig:forest-structure-plot-level-scatter}) highlight how relationships between Cover and Height with AGB, BA, and QMD are much more variable in ground-truth plots than are represented in the training tiles (where AGB, BA, and QMD are predicted instead of directly observed). This divergence also highlights that the performance metrics from training tiles for target attributes other than those directly sensed by lidar (i.e., Cover, Height) should not be misinterpreted as indicators of accuracy or precision for AGB, BA, or QMD predictions in real-world applications.

\subsection{Evaluation on Pacific Northwest Field Plots}\label{sec:plot-scale-results}

\paragraph{Allometric Forest Structure.}

\begin{figure}[htbp]
\centering
\includegraphics[width=\textwidth]{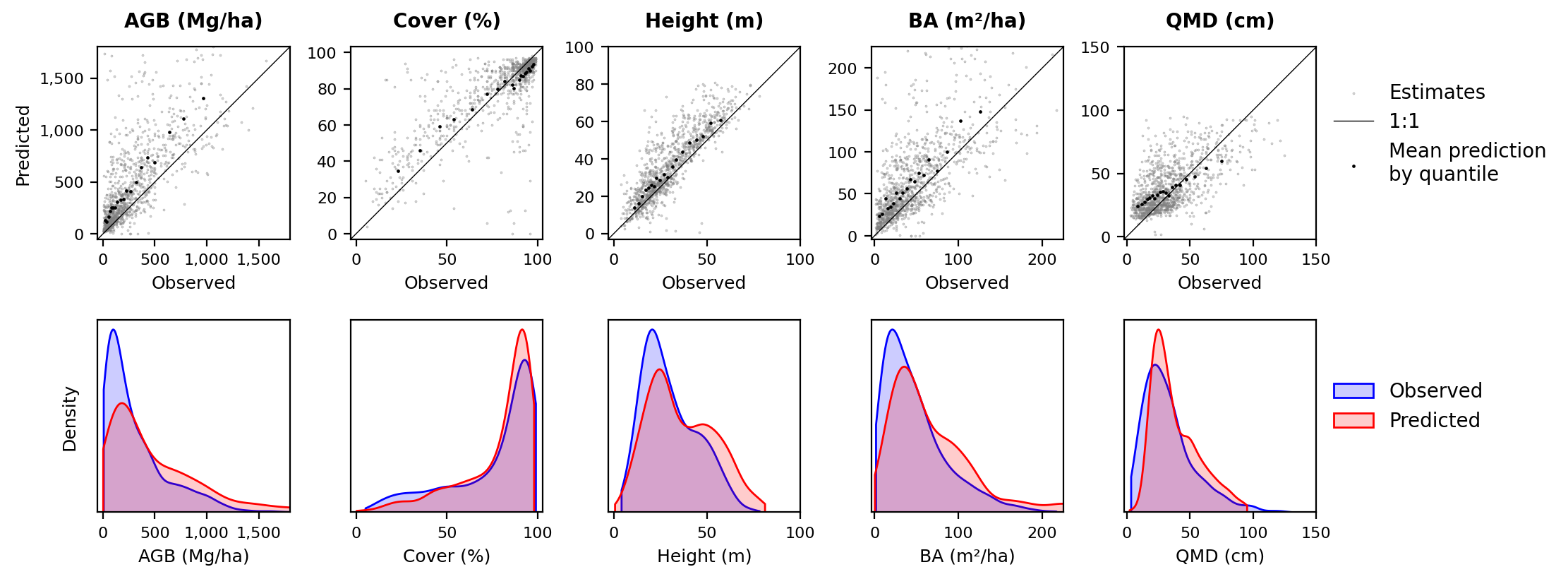}
\caption{Plot-level evaluation of lidar-derived training tiles against independent Pacific Northwest field plots not used in allometric model fitting (n=1,086). Scatter plots (top) show the values in training tiles (``Predicted'') where co-located field inventory plots (``Observed'') exist. The black points show mean predicted values for each quantile of observations (5th-95th percentile, step=5). Density plots (bottom) compare the distributions of predicted and observed values across all plots. Slight overprediction biases for AGB, Height, and BA represent a source of systematic error inherited by the satellite model from the allometric model, since these tiles serve as training targets.}
\label{fig:allometric-structure-plot-level-scatter}
\end{figure}

Error in the underlying allometric model employed to generate the training tiles propagates into the satellite model. When we evaluated how well the training tiles compared to the independent field inventory plots which were not used to fit the allometric model (see Figure \ref{fig:allometric-structure-plot-level-scatter}), we observed slight overprediction biases for AGB, Height, and BA. We observed tighter correspondence between QMD for field observations and training tiles than we did for predictions by the satellite model against these field plots, but the correlation was still the weakest among all the target variables.

\paragraph{Satellite Forest Structure.}

\begin{figure}[htbp]
\centering
\includegraphics[width=\textwidth]{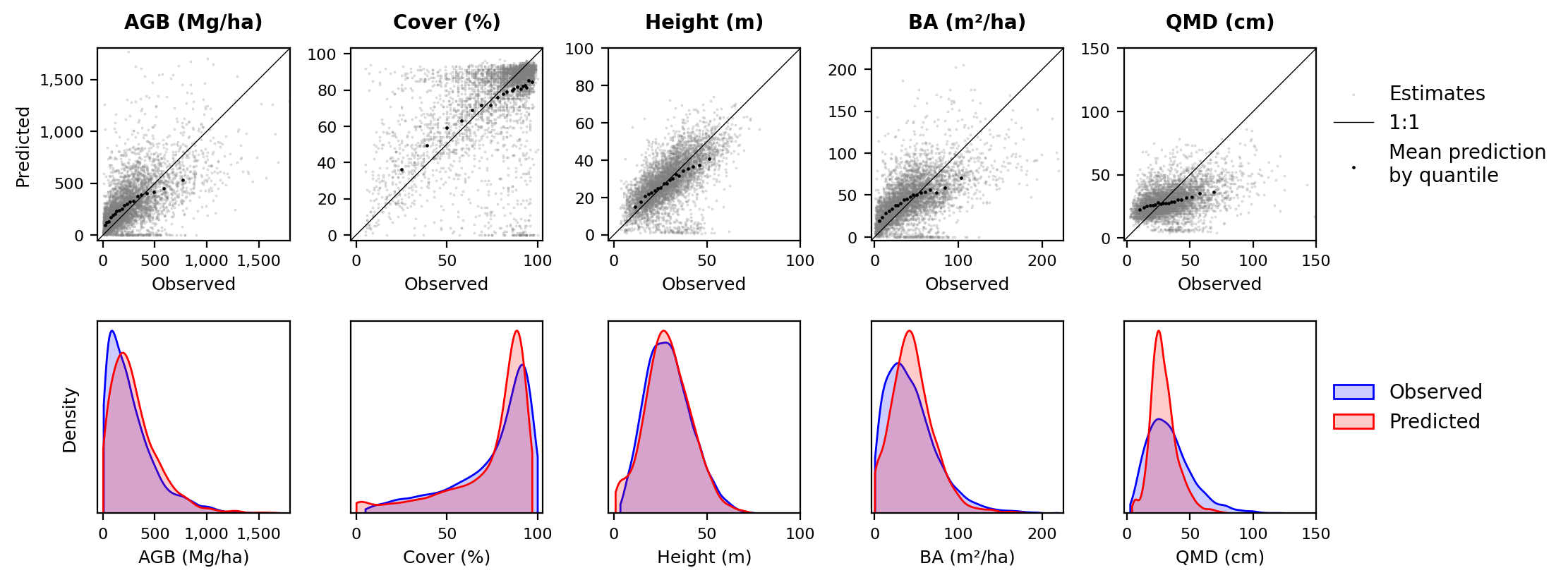}
\caption{Overall performance of satellite forest structure model on independent field observations (n=4,538). The black points in each scatter plot show the average value predicted for each quantile of observations, spanning the 5th through 95th percentiles in steps of 5.}
\label{fig:forest-structure-plot-level-scatter}
\end{figure}

We found strong correspondence between observed and predicted values for AGB, Cover, Height, and BA with modest saturation effects beginning to emerge around 450 Mg/ha for AGB, 40m for height, and 80\% canopy cover (see Figure \ref{fig:forest-structure-plot-level-scatter} and Table \ref{tab:forest-structure-plot-table}). The density curves of predictions and field plot observations show that satellite model-generated predictions spanning the full range of observed values, including very high biomass and very tall canopy conditions.

The effects of the time lag between field observations (2010-2018) and forest structure predictions (2024) were evident in many plots that are likely to have been disturbed since the time of field measurement. These plots are visible along the x-axis in the top row of graphs in Figure \ref{fig:forest-structure-plot-level-scatter} where near-zero predictions were made in 2024 but where observed forest attributes from 2010-2018 were substantially larger.

The satellite forest structure model showed limited ability to discern the variation in average tree diameter (QMD). The density plot for QMD showed model predictions biased toward the mean with a distribution compressed toward the mean of the distribution of observed QMD. This pattern becomes even more pronounced when BA and QMD are combined to derive TPH, indicating that the satellite forest structure model is not able to precisely detect tree diameters and stem densities across landscapes to the same degree that is apparent for AGB, BA, Height, and Cover predictions.

Summary statistics (see Table \ref{tab:forest-structure-plot-table}) confirmed limited bias (averaging less than 10\% of mean observed values for AGB, 5\% of mean observed values for BA, and 1\% of observed mean values for Cover and Height) and modest correlations between observations and predictions. $R^2$ was low across predicted structure attributes, highlighting the amount of noise and error involved with plot-level comparisons even when reasonable correlations between observations and predictions (Pearson's $r$) were found.

\begin{table}[htbp]
\centering
\caption{Summary statistics for plot-scale evaluation of satellite forest structure estimates 2024 against field plots (2010-2018). Obs and Pred columns display the mean ± 1 standard deviation for Observed and Predicted values, respectively.}
\label{tab:forest-structure-plot-table}
\begin{tabular}{lccccccc}
\toprule
Target & MAE & RMSE & Mean Bias & $R^2$ & Pearson's $r$ & Obs & Pred \\
\midrule
AGB (Mg/ha) & 149.9 & 218.1 & +37.6 & 0.17 & 0.60 & 270 ± 239 & 307 ± 243 \\
Cover (\%) & 13.4 & 22.0 & -1.5 & 0.02 & 0.54 & 75 ± 22 & 74 ± 23 \\
Height (m) & 7.6 & 10.6 & -0.3 & 0.25 & 0.63 & 29 ± 12 & 29 ± 12 \\
BA (m²/ha) & 19.9 & 27.4 & +1.8 & 0.24 & 0.57 & 44 ± 32 & 46 ± 27 \\
QMD (cm) & 12.8 & 17.7 & -4.8 & 0.10 & 0.43 & 34 ± 19 & 29 ± 11 \\
\bottomrule
\end{tabular}
\end{table}

\subsection{Regional Distributions}\label{sec:regional-results}

\begin{figure}[htbp]
\centering
\includegraphics[width=\textwidth]{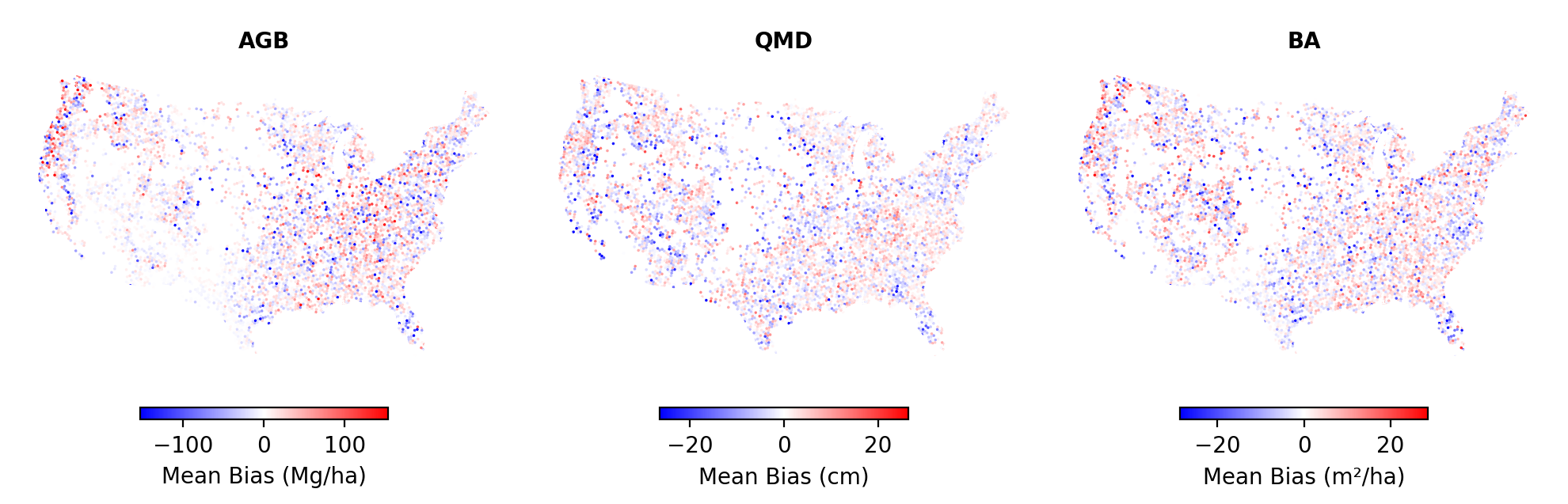}
\caption{Geographic distribution of allometric model bias on FIA test-partition subplots evaluated at the scale of 64,000 hectare hexagons (n=6,643). Red hexagons indicate overprediction, while blue indicate underprediction. The absence of obvious geographic clustering suggests the allometric model does not exhibit systematic regional biases.}
\label{fig:allometric-structure-hex-bias}
\end{figure}

\paragraph{Allometric Forest Structure.}
We aggregated residuals for predictions on FIA subplots within the test partition to the scale of 64,000ha hexagons for visualization of the spatial variation in model bias (see Figure \ref{fig:allometric-structure-hex-bias}). We did not observe any substantial geographic concentrations of over- or under-prediction biases for any of the allometric model's targets, but rather observed a mixture of over- and under-predictions across the hexagons.

\begin{figure}[htbp]
\centering
\includegraphics[width=\textwidth]{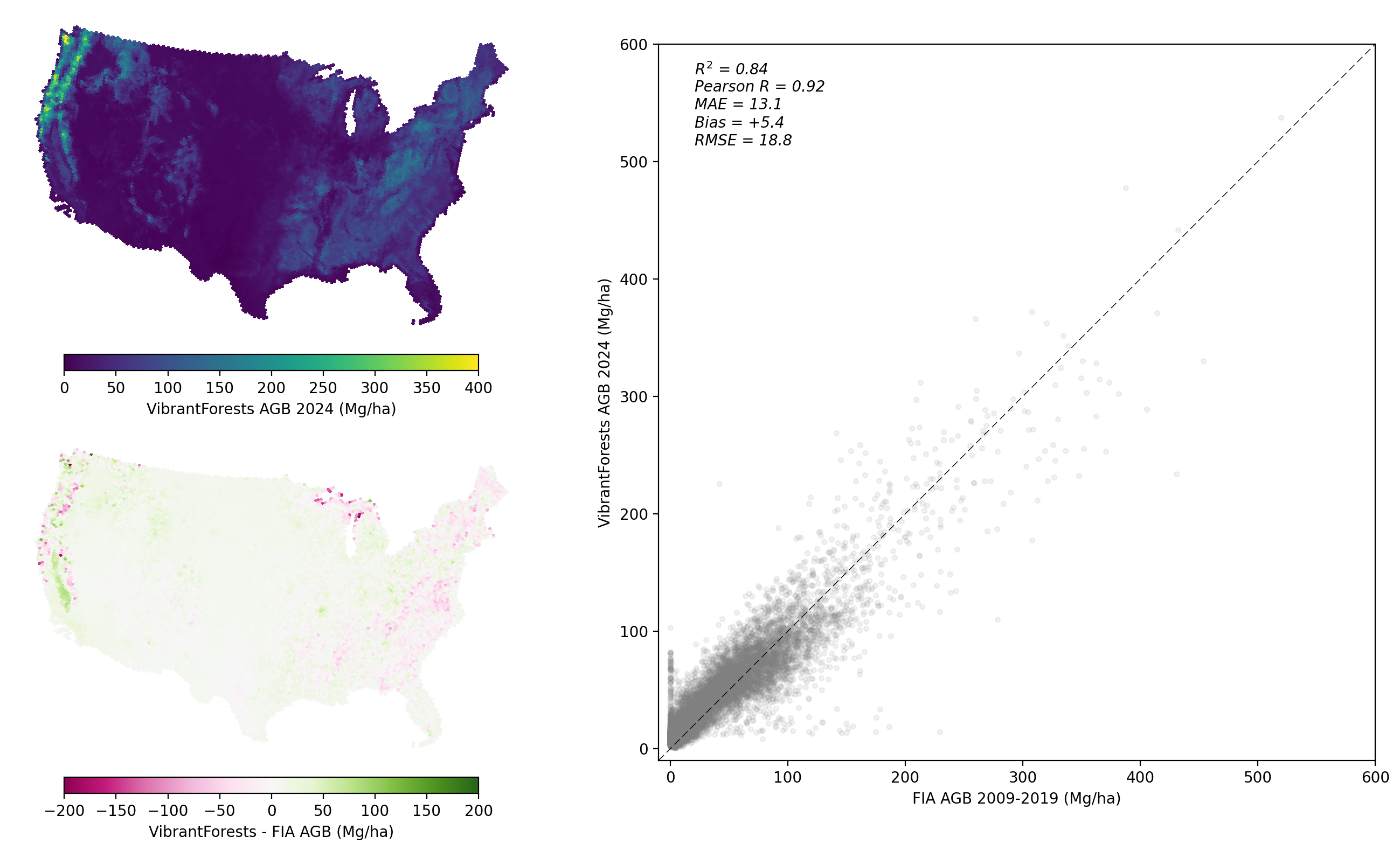}
\caption{Comparison of satellite based AGB predictions (2024) with hexagon-level FIA observations from \cite{menlove-2020} (2009-2019). Top left: wall-to-wall AGB predictions across CONUS. Bottom left: difference map (VibrantForests minus FIA), with green indicating overprediction and magenta underprediction. Right: scatter plot of hexagon-level averages (summary statistics inset). The strong agreement ($R^2$=0.84, Pearson's $r$=0.92) across the full range of observed values demonstrates low regional bias, with the exception of modest overprediction in the Central Valley of California and underprediction along the Appalachians, and edge effects apparent along the land border with water among the Lake States.}
\label{fig:conus-agb-vs-menlove}
\end{figure}

\begin{figure}[htbp]
\centering
\includegraphics[width=\textwidth]{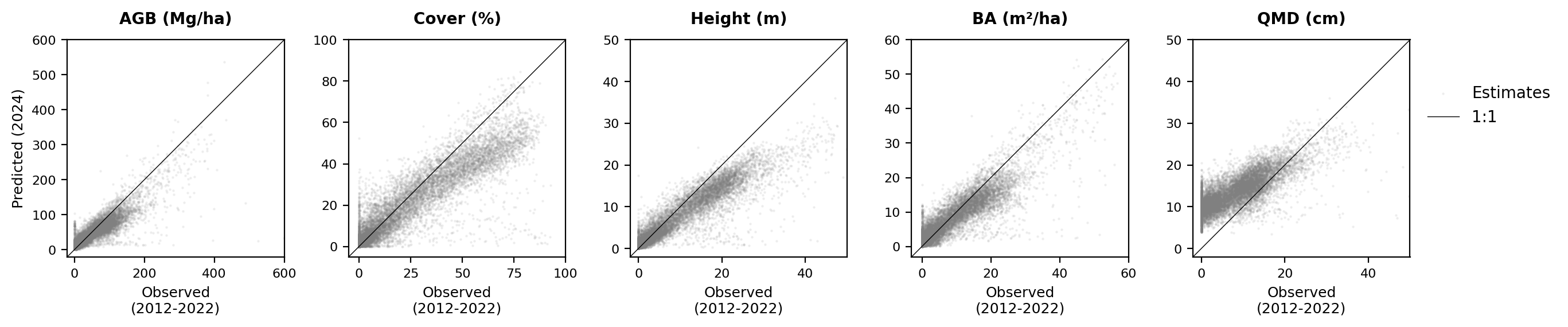}
\caption{Hexagon-level (64,000 ha) comparisons between satellite-based (2024) predictions and FIA observations (2012–2022) for all five target variables (n=12,604 hexagons). Points near the 1:1 line indicate low regional bias. QMD is a notable exception, with predictions compressed toward higher values relative to the observed range, consistent with the model's limited sensitivity to stem diameter discussed in Section \ref{sec:plot-scale-results}.}
\label{fig:conus-hexagons}
\end{figure}

\paragraph{Satellite Forest Structure.}
Regional evaluations against FIA observations at the scale of 64,000ha hexagons provided additional evidence of low bias (see Figures \ref{fig:conus-agb-vs-menlove} and \ref{fig:conus-hexagons}), and revealed limited geographic concentrations of bias, suggesting no systematic regional deficiencies in model reliability. We noticed a modest tendency to overpredict AGB in the Central Valley of southern California and a smaller underprediction bias for AGB along the Appalachian Mountains of the eastern USA. We suspect the underprediction biases observed along the northern borders of Michigan and Minnesota are artifacts of hexagons that are dominated by water because high proportions of water cover result in lower model predictions when summarizing to the hexagon level.

All other target variables had a similar pattern to what was observed for AGB. We noticed a tendency for the model to produce Cover and Height averages at regional scales that were modestly lower than the FIA observations (see Table \ref{tab:forest-structure-region-table}). Similar to the plot-scale evaluations, we noticed QMD predictions were relatively less sensitive and precise.

\begin{table}[htbp]
\centering
\caption{Summary statistics for region-scale evaluation of satellite based estimates for 2024 against FIA (2012-2022). Obs and Pred columns depict the mean ± standard deviation for the observed and predicted attributes, respectively.}
\label{tab:forest-structure-region-table}
\begin{tabular}{lccccccc}
\toprule
Target & MAE & RMSE & Mean Bias & $R^2$ & Pearson's $r$ & Obs & Pred \\
\midrule
AGB (Mg/ha) & 15.9 & 26.6 & -2.37 & 0.78 & 0.90 & 41.6 ± 56.4 & 39.2 ± 42.1 \\
Cover (\%) & 7.1 & 11.5 & -3.90 & 0.77 & 0.91 & 22.3 ± 24.1 & 18.4 ± 18.0 \\
Height (m) & 2.7 & 4.6 & -1.85 & 0.76 & 0.93 & 8.4 ± 9.5 & 6.6 ± 6.5 \\
BA (m²/ha) & 2.8 & 4.5 & -0.54 & 0.77 & 0.89 & 8.0 ± 9.4 & 7.4 ± 7.2 \\
QMD (cm) & 6.6 & 7.6 & +5.91 & -0.13 & 0.76 & 7.4 ± 7.2 & 13.3 ± 4.2\\
\bottomrule
\end{tabular}
\end{table}

\section{Discussion}\label{sec:discussion}
Despite the fact that we expect some bias to be inherited by the satellite model from the upstream allometric model, the patterns of biases that were observed when evaluating the allometric model predictions against the FIA test partition were not observed when satellite predictions were evaluated against the lidar training tiles nor against confidently-located field plots in the Pacific Northwest. 

We suspect the difference in how canopy cover and height are derived from the FIA plots versus lidar data contribute to this shift. Regardless of the mechanism, the outcome was favorable: the saturation pattern for AGB observed in the allometric model at 150-250 Mg/ha (see Figure \ref{fig:allometric-structure-subplot-level-scatter}) appears to be have been pushed substantially outward to 450-500 Mg/ha in both the lidar-derived training tiles (see Figure \ref{fig:allometric-structure-plot-level-scatter}) and satellite predictions (see Figure \ref{fig:forest-structure-plot-level-scatter}) when evaluated against the Pacific Northwest field inventory plots.

As these models are applied to increasingly larger spatial extents, the limited prediction bias translates into very strong correspondence for the satellite based predictions with observations from the FIA program summarized at landscape scales (see Figures \ref{fig:conus-agb-vs-menlove} and \ref{fig:conus-hexagons} and Table \ref{tab:forest-structure-region-table}). When evaluation is focused on finer spatial scales (e.g., plot-level), the variance among samples of forest structure targets increases substantially (see Table \ref{tab:forest-structure-plot-table} and Figure \ref{fig:forest-structure-plot-level-scatter}) and the amount of variation that can be explained by model predictions declines. 

In contrast to other computer vision models that have been trained on higher-resolution imagery to predict canopy height and cover (e.g., \cite{chang-2025, anderson-2026, tolan-2024, liu-2023}), we noticed relatively limited height saturation or under-prediction by our satellite model in tall canopies. While \cite{chang-2025, anderson-2026, tolan-2024, liu-2023} report under-estimation bias beginning in canopy heights in the 20-30m range, evaluations of our satellite forest structure model against the field-observed canopy heights in the Pacific Northwest field plots showed unbiased predictions into canopies as tall as 40m, with average under-prediction bias above this range growing to 5-10m in canopies observed to be 50m tall.

The ability for our satellite forest structure model to make relatively unbiased predictions in very dense and tall canopies translates into improved sensitivity of AGB predictions in very high biomass stands. We encountered numerous instances where the satellite forest structure predicted AGB estimates above 1,000 Mg/ha. When these extremely high AGB values were initially encountered during preparation of these data for use within the Vibrant Planet Platform, these pixels were initially flagged as potential outliers during Quality Assessment. However, upon further investigation these predictions were found to correspond to groves and reserves of old growth redwoods found in well-known campgrounds and recreation maps.

In addition to showing strong predictive performance in tall/dense canopies, we also observed that the satellite forest structure model was capable of making predictions at or near zero in areas that had been disturbed, as represented by the points along the x-axis in Figure \ref{fig:forest-structure-plot-level-scatter} where predictions are near zero but observations were substantially greater than zero at the time of field observation. Although the satellite forest structure model was not explicitly trained to detect disturbance or to produce time series estimates, its sensitivity to these occurrences is promising for future work, and has been reinforced by recent observations in operational use with the Vibrant Planet Platform in fire-scars and harvested areas.

In general, we found the satellite forest structure model was able to predict core forest structure attributes with the quality needed for wildfire risk assessment and forest restoration planning at landscape scale. Although the satellite forest structure model's predictions of QMD and TPH showed less sensitivity and explanatory power, our experience with landscape-scale planning and prioritization indicates precise estimates of these attributes are not critical. Instead, we more often observe attributes like basal area, total volume or biomass, canopy cover, and canopy height and other derived metrics like Stand Density Index or merchantable timber volume (alongside indicators of wildfire risk and mitigation potential) given stronger interest by managers and decision-makers conducting landscape-scale planning. 

\section{Conclusion}\label{sec:conclusion}
We introduced the VibrantForests framework and its forest structure model components being applied for multi-target quantile regression to predict canopy cover, canopy height, aboveground biomass, basal area, and quadratic mean diameter at 10m resolution from Sentinel-2 imagery. Our training data generation process incorporated direct lidar-detected features augmented by allometrically-estimated forest structure attributes (AGB, BA, QMD) to allow a computer vision model to leverage spatial context in multi-target predictions. Modeling choices include the adoption of a multi-target quantile regression loss function contributed to strong predictive performance by the satellite forest structure model extending further into the range of high-biomass and tall-canopy conditions than other satellite-based models. The satellite based model showed strong predictive ability in recovering canopy cover, height, AGB, and basal area, but it showed more limited ability to recover the drivers of variation in quadratic mean diameter. At regional scales, we observed minimal systematic bias with no significant geographic clustering of over- or under-prediction biases in any forested ecoregions.

The satellite forest structure model helps address a key limitation in large-area forest and wildfire planning by delivering wall-to-wall estimates at an annual cadence and 10m resolution. These outputs serve as a backbone from which coherent downstream applications can be derived, including forest product estimation, stocking level assessments, and fuel characterization, while reducing unintended artifacts that typically emerge when data sources with varying purposes, vintages, and prediction quality are combined. VibrantForests is now being applied to equip practitioners with the reliable wall-to-wall and up-to-date information needed to make data-informed decisions about forest management and wildfire resilience across dynamic landscapes. 

\section{Acknowledgments}\label{sec:acknowledgments}
The development of these forest structure models would not have been possible without the ability to build upon a large collection of lidar data acquired and processed through contributions by many colleagues at Vibrant Planet, including Mitchell Gritts, Zoe Statman-Weil, Ryan Herring, Paige Maas, Bogdan State, Danielle Perrot, Colton Miller, Janet Wiener, and Tory Nelson.

The research and development described here was supported by grant funding from the Doris Duke Foundation to American Forests for the development of the \href{https://forestinnovationplatform.org/}{Forest Innovation Platform}.

This material is based upon work supported by the U.S. Department of Agriculture, under agreement number NR233A750004G042, and by the USDA Forest Service, under agreement number \#24-CA-11132544-064. Any opinions, findings, conclusions, or recommendations expressed in this publication are those of the author(s) and do not necessarily reflect the views of the U.S. Department of Agriculture. In addition, any reference to specific brands or types of products or services does not constitute or imply an endorsement by the U.S. Department of Agriculture for those products or services.

\section{Author Contributions}\label{sec:contributions}
Conceptualization: L.J.Z., D.D.D., T.C., and N.E.R.; 
Methodology: L.J.Z., D.D.D, V.A.L, and T.C.; 
Data curation: L.J.Z., D.D.D., V.A.L., T.C., and K.N.; 
Software: L.J.Z., D.D.D., and V.A.L; 
Visualization: D.D.D. and L.J.Z; 
Writing-original draft: D.D.D., L.J.Z., and C.WS.; 
Writing—review and editing: D.D.D., L.J.Z., C.WS., V.A.L., T.C., K.N., A.G., and N.E.R.; 
Supervision: T.C., A.G., and G.B.; 
Project administration: T.C., K.A.D., and N.E.R.; and 
Funding acquisition: D.D.D., K.A.D., and S.C.

\newpage
\section*{Data Availability}
\addcontentsline{toc}{section}{Data Availability}

Several raster outputs from VibrantForests based on imagery year 2024 have been packaged for public release in the forthcoming Forest Innovation Platform. As we further pursue journal publication, we also expect to release a broader collection of raster data generated by VibrantForests through the Vibrant Planet Data Commons (www.vpdatacommons.org).

\newpage
\bibliography{references}

@misc{falcon-2020,
	title = {{PyTorchLightning}/pytorch-lightning: 0.7.6 release},
	copyright = {Open Access},
	shorttitle = {{PyTorchLightning}/pytorch-lightning},
	url = {https://zenodo.org/record/3828935},
	doi = {10.5281/ZENODO.3828935},
	urldate = {2026-06-17},
	publisher = {Zenodo},
	author = {Falcon, William and Borovec, Jirka and Wälchli, Adrian and Eggert, Nic and Schock, Justus and Jordan, Jeremy and Skafte, Nicki and {Ir1dXD} and Bereznyuk, Vadim and Harris, Ethan and {Tullie Murrell} and Yu, Peter and Præsius, Sebastian and Addair, Travis and Zhong, Jacob and Lipin, Dmitry and Uchida, So and {Shreyas Bapat} and Schröter, Hendrik and Dayma, Boris and Karnachev, Alexey and {Akshay Kulkarni} and {Shunta Komatsu} and {Martin.B} and {Jean-Baptiste SCHIRATTI} and Mary, Hadrien and Byrne, Donal and {Cristobal Eyzaguirre} and {Cinjon} and Bakhtin, Anton},
	month = may,
	year = {2020},
}

@article{crookston-2005,
	title = {The {Forest} {Vegetation} {Simulator}: {A} review of its structure, content, and applications},
	volume = {49},
	shorttitle = {The {Forest} {Vegetation} {Simulator}},
	url = {https://research.fs.usda.gov/treesearch/28474},
	doi = {10.1016/j.compag.2005.02.003},
	language = {en},
	number = {1},
	urldate = {2026-06-01},
	journal = {Computers and Electronics in Agriculture},
	author = {Crookston, Nicholas L. and Dixon, Gary E.},
	year = {2005},
	pages = {60--80},
}

@inproceedings{loshchilov2018,
	title = {Decoupled {Weight} {Decay} {Regularization}},
	url = {https://openreview.net/forum?id=Bkg6RiCqY7},
	language = {en},
	urldate = {2026-05-23},
	booktitle = {{ICLR} 2019},
	author = {Loshchilov, Ilya and Hutter, Frank},
	month = sep,
	year = {2018},
}

@inproceedings{he2022,
	address = {New Orleans, LA, USA},
	title = {Masked {Autoencoders} {Are} {Scalable} {Vision} {Learners}},
	copyright = {https://doi.org/10.15223/policy-029},
	isbn = {978-1-6654-6946-3},
	url = {https://ieeexplore.ieee.org/document/9879206/},
	doi = {10.1109/CVPR52688.2022.01553},
	language = {en},
	urldate = {2026-05-23},
	booktitle = {2022 {IEEE}/{CVF} {Conference} on {Computer} {Vision} and {Pattern} {Recognition} ({CVPR})},
	publisher = {IEEE},
	author = {He, Kaiming and Chen, Xinlei and Xie, Saining and Li, Yanghao and Dollar, Piotr and Girshick, Ross},
	month = jun,
	year = {2022},
	pages = {15979--15988},
}

@inproceedings{lin2017,
	address = {Honolulu, HI},
	title = {Feature {Pyramid} {Networks} for {Object} {Detection}},
	isbn = {978-1-5386-0457-1},
	url = {http://ieeexplore.ieee.org/document/8099589/},
	doi = {10.1109/CVPR.2017.106},
	language = {en},
	urldate = {2026-05-23},
	booktitle = {2017 {IEEE} {Conference} on {Computer} {Vision} and {Pattern} {Recognition} ({CVPR})},
	publisher = {IEEE},
	author = {Lin, Tsung-Yi and Dollar, Piotr and Girshick, Ross and He, Kaiming and Hariharan, Bharath and Belongie, Serge},
	month = jul,
	year = {2017},
	pages = {936--944},
}

@article{pedregosa2011,
	title = {Scikit-learn: {Machine} {Learning} in {Python}},
	volume = {12},
	issn = {1532-4435},
	shorttitle = {Scikit-learn},
	url = {https://dl.acm.org/doi/10.5555/1953048.2078195},
	number = {null},
	urldate = {2026-05-23},
	journal = {The Journal of Machine Learning Research},
	author = {Pedregosa, Fabian and Varoquaux, Gaël and Gramfort, Alexandre and Michel, Vincent and Thirion, Bertrand and Grisel, Olivier and Blondel, Mathieu and Prettenhofer, Peter and Weiss, Ron and Dubourg, Vincent and Vanderplas, Jake and Passos, Alexandre and Cournapeau, David and Brucher, Matthieu and Perrot, Matthieu and Duchesnay, Édouard},
	month = nov,
	year = {2011},
	pages = {2825--2830},
}

@techreport{lapuma-2023,
	address = {Reston, VA},
	type = {Open-{File} {Report}},
	title = {{LANDFIRE} {Technical} {Documentation}},
	issn = {ISSN 2331-1258},
	url = {https://doi.org/10.3133/ofr20231045},
	doi = {10.3133/ofr20231045},
	language = {en},
	number = {2023-1045},
	institution = {U.S. Department of Interior, U.S. Geological Survey},
	author = {La Puma, Inga P.},
	year = {2023},
	pages = {103},
}

@techreport{-p,
	address = {Boston, MA},
	type = {Data {Specifications}},
	title = {Above-{Ground} {Biomass} ({AGB}) {Stock} and {Change} {Product}},
	url = {https://static1.squarespace.com/static/65ae8dc337ea664cbf34181c/t/6981db51c8bb6f5126514640/1770117969479/Data+specifications_Chloris+Geospatial_2026.pdf},
	urldate = {2026-03-19},
	institution = {Chloris Geospatial},
	author = {{Chloris Geospatial}},
	year = {2026},
	pages = {2},
}

@misc{butler-2024,
	title = {{PDAL}/{PDAL}: 2.8.3},
	copyright = {Creative Commons Attribution 4.0 International},
	shorttitle = {{PDAL}/{PDAL}},
	url = {https://zenodo.org/doi/10.5281/zenodo.2616780},
	doi = {10.5281/ZENODO.2616780},
	urldate = {2026-05-13},
	publisher = {Zenodo},
	author = {Butler, Howard and Bell, Andrew and Gerlek, Michael P. and {chambbj} and Gadomski, Pete and Manning, Connor and Łoskot, Mateusz and Couwenberg, Bas and Barker, Norman and Ramsey, Paul and Rouault, Even and Dark, Julia and Mann, Kyle and {Guilhem Villemin} and Chaulet, Nicolas and Foster, Claire and Rosen, Michael and Moore, Ogi and Lewis, Scott and {Grigory} and McKelvey, Kirk and Dobias, Martin and Bell, Isaac and Smith, Michael D. and {Bram} and {xantares} and Vergara, Vicky and Coup, Robert and Evers, Kristian},
	month = dec,
	year = {2024},
}

@article{butler-2021,
	title = {{PDAL}: {An} open source library for the processing and analysis of point clouds},
	volume = {148},
	issn = {00983004},
	shorttitle = {{PDAL}},
	url = {https://linkinghub.elsevier.com/retrieve/pii/S0098300420306518},
	doi = {10.1016/j.cageo.2020.104680},
	language = {en},
	urldate = {2026-05-13},
	journal = {Computers \& Geosciences},
	author = {Butler, Howard and Chambers, Bradley and Hartzell, Preston and Glennie, Craig},
	month = mar,
	year = {2021},
	pages = {104680},
}

@misc{usgs-2026,
	type = {Point {Cloud}},
	title = {{USGS} {3DEP} {LiDAR} {Point} {Clouds}},
	url = {https://registry.opendata.aws/usgs-lidar/},
	urldate = {2026-05-13},
	publisher = {Registry of Open Data on AWS},
	author = {{USGS}},
	year = {2026},
}

@misc{usepa-2015,
	address = {Corvallis, OR},
	type = {Data and {Tools}},
	title = {Level {III} and {IV} {Ecoregions} of the {Continental} {United} {States}},
	url = {https://www.epa.gov/eco-research/level-iii-and-iv-ecoregions-continental-united-states},
	language = {en},
	urldate = {2026-05-13},
	journal = {United States Environmental Protection Agency},
	publisher = {US EPA},
	author = {{US EPA}},
	month = nov,
	year = {2015},
}

@article{anderson-2026,
	title = {Forest {Carbon} {Diligence}: {Digital} {MRV} for mapping forest structure and carbon stocks},
	volume = {612},
	issn = {0378-1127},
	shorttitle = {Forest {Carbon} {Diligence}},
	url = {https://www.sciencedirect.com/science/article/pii/S0378112726002070},
	doi = {10.1016/j.foreco.2026.123709},
	urldate = {2026-05-13},
	journal = {Forest Ecology and Management},
	author = {Anderson, Christopher B. and Joseph, Maxwell B. and Söthe, Camile and de Souza Mendes, Flávia and Maschler, Thomas and McCarthy, Ryan C. and Mascaro, Joseph and O’Shea, Tara and Rosenthal, Amy and Marvin, David C.},
	month = jul,
	year = {2026},
	pages = {123709},
}

@article{dubayah-2020,
	title = {The {Global} {Ecosystem} {Dynamics} {Investigation}: {High}-resolution laser ranging of the {Earth}’s forests and topography},
	volume = {1},
	issn = {2666-0172},
	shorttitle = {The {Global} {Ecosystem} {Dynamics} {Investigation}},
	url = {https://www.sciencedirect.com/science/article/pii/S2666017220300018},
	doi = {10.1016/j.srs.2020.100002},
	urldate = {2026-05-13},
	journal = {Science of Remote Sensing},
	author = {Dubayah, Ralph and Blair, James Bryan and Goetz, Scott and Fatoyinbo, Lola and Hansen, Matthew and Healey, Sean and Hofton, Michelle and Hurtt, George and Kellner, James and Luthcke, Scott and Armston, John and Tang, Hao and Duncanson, Laura and Hancock, Steven and Jantz, Patrick and Marselis, Suzanne and Patterson, Paul L. and Qi, Wenlu and Silva, Carlos},
	month = jun,
	year = {2020},
	pages = {100002},
}

@article{wulder-2022,
	title = {Fifty years of {Landsat} science and impacts},
	volume = {280},
	issn = {0034-4257},
	url = {https://www.sciencedirect.com/science/article/pii/S0034425722003054},
	doi = {10.1016/j.rse.2022.113195},
	urldate = {2026-05-13},
	journal = {Remote Sensing of Environment},
	author = {Wulder, Michael A. and Roy, David P. and Radeloff, Volker C. and Loveland, Thomas R. and Anderson, Martha C. and Johnson, David M. and Healey, Sean and Zhu, Zhe and Scambos, Theodore A. and Pahlevan, Nima and Hansen, Matthew and Gorelick, Noel and Crawford, Christopher J. and Masek, Jeffrey G. and Hermosilla, Txomin and White, Joanne C. and Belward, Alan S. and Schaaf, Crystal and Woodcock, Curtis E. and Huntington, Justin L. and Lymburner, Leo and Hostert, Patrick and Gao, Feng and Lyapustin, Alexei and Pekel, Jean-Francois and Strobl, Peter and Cook, Bruce D.},
	month = oct,
	year = {2022},
	pages = {113195},
}

@phdthesis{diaz2024,
	address = {Seattle, WA},
	type = {Ph.{D}. {Dissertation}},
	title = {Leveraging {Open} {Data} to {Support} {Forest} {Mapping}, {Modeling}, and {Policy} {Analysis} in the {Pacific} {Northwest}, {USA}},
	url = {http://hdl.handle.net/1773/51176},
	language = {en\_US},
	urldate = {2026-05-09},
	school = {University of Washington},
	author = {Diaz, David Daniel},
	month = feb,
	year = {2024},
}

@misc{-2026a,
	title = {Earth {Search} {STAC} {API}},
	url = {https://github.com/Element84/earth-search},
	urldate = {2026-04-14},
	publisher = {Element 84},
	author = {{Element84}},
	month = apr,
	year = {2026},
	note = {original-date: 2023-05-09T02:21:20Z},
}

@article{drusch-2012,
	series = {The {Sentinel} {Missions} - {New} {Opportunities} for {Science}},
	title = {Sentinel-2: {ESA}'s {Optical} {High}-{Resolution} {Mission} for {GMES} {Operational} {Services}},
	volume = {120},
	issn = {0034-4257},
	shorttitle = {Sentinel-2},
	url = {https://www.sciencedirect.com/science/article/pii/S0034425712000636},
	doi = {10.1016/j.rse.2011.11.026},
	urldate = {2026-04-14},
	journal = {Remote Sensing of Environment},
	author = {Drusch, M. and Del Bello, U. and Carlier, S. and Colin, O. and Fernandez, V. and Gascon, F. and Hoersch, B. and Isola, C. and Laberinti, P. and Martimort, P. and Meygret, A. and Spoto, F. and Sy, O. and Marchese, F. and Bargellini, P.},
	month = may,
	year = {2012},
	pages = {25--36},
}

@techreport{-n,
	title = {Confronting the {Wildfire} {Crisis}: {Initial} {Landscape} {Investments} to {Protect} {Communities} and {Improve} {Resilience} in {America}'s {Forests}},
	url = {https://www.fs.usda.gov/sites/default/files/fs_media/fs_document/WCS-Initial-Landscapes.pdf},
	number = {FS-1187d},
	urldate = {2026-04-14},
	institution = {U.S. Department of Agriculture, Forest Service},
	author = {{USFS}},
	month = apr,
	year = {2022},
	pages = {17},
}

@article{liu-2023,
	title = {The overlooked contribution of trees outside forests to tree cover and woody biomass across {Europe}},
	volume = {9},
	url = {https://www.science.org/doi/10.1126/sciadv.adh4097},
	doi = {10.1126/sciadv.adh4097},
	number = {37},
	urldate = {2026-04-14},
	journal = {Science Advances},
	publisher = {American Association for the Advancement of Science},
	author = {Liu, Siyu and Brandt, Martin and Nord-Larsen, Thomas and Chave, Jerome and Reiner, Florian and Lang, Nico and Tong, Xiaoye and Ciais, Philippe and Igel, Christian and Pascual, Adrian and Guerra-Hernandez, Juan and Li, Sizhuo and Mugabowindekwe, Maurice and Saatchi, Sassan and Yue, Yuemin and Chen, Zhengchao and Fensholt, Rasmus},
	month = sep,
	year = {2023},
	pages = {eadh4097},
}

@article{duncanson-2022,
	title = {Aboveground biomass density models for {NASA}’s {Global} {Ecosystem} {Dynamics} {Investigation} ({GEDI}) lidar mission},
	volume = {270},
	issn = {0034-4257},
	url = {https://www.sciencedirect.com/science/article/pii/S0034425721005654},
	doi = {10.1016/j.rse.2021.112845},
	urldate = {2026-04-10},
	journal = {Remote Sensing of Environment},
	author = {Duncanson, Laura and Kellner, James R. and Armston, John and Dubayah, Ralph and Minor, David M. and Hancock, Steven and Healey, Sean P. and Patterson, Paul L. and Saarela, Svetlana and Marselis, Suzanne and Silva, Carlos E. and Bruening, Jamis and Goetz, Scott J. and Tang, Hao and Hofton, Michelle and Blair, Bryan and Luthcke, Scott and Fatoyinbo, Lola and Abernethy, Katharine and Alonso, Alfonso and Andersen, Hans-Erik and Aplin, Paul and Baker, Timothy R. and Barbier, Nicolas and Bastin, Jean Francois and Biber, Peter and Boeckx, Pascal and Bogaert, Jan and Boschetti, Luigi and Boucher, Peter Brehm and Boyd, Doreen S. and Burslem, David F. R. P. and Calvo-Rodriguez, Sofia and Chave, Jérôme and Chazdon, Robin L. and Clark, David B. and Clark, Deborah A. and Cohen, Warren B. and Coomes, David A. and Corona, Piermaria and Cushman, K. C. and Cutler, Mark E. J. and Dalling, James W. and Dalponte, Michele and Dash, Jonathan and de-Miguel, Sergio and Deng, Songqiu and Ellis, Peter Woods and Erasmus, Barend and Fekety, Patrick A. and Fernandez-Landa, Alfredo and Ferraz, Antonio and Fischer, Rico and Fisher, Adrian G. and García-Abril, Antonio and Gobakken, Terje and Hacker, Jorg M. and Heurich, Marco and Hill, Ross A. and Hopkinson, Chris and Huang, Huabing and Hubbell, Stephen P. and Hudak, Andrew T. and Huth, Andreas and Imbach, Benedikt and Jeffery, Kathryn J. and Katoh, Masato and Kearsley, Elizabeth and Kenfack, David and Kljun, Natascha and Knapp, Nikolai and Král, Kamil and Krůček, Martin and Labrière, Nicolas and Lewis, Simon L. and Longo, Marcos and Lucas, Richard M. and Main, Russell and Manzanera, Jose A. and Martínez, Rodolfo Vásquez and Mathieu, Renaud and Memiaghe, Herve and Meyer, Victoria and Mendoza, Abel Monteagudo and Monerris, Alessandra and Montesano, Paul and Morsdorf, Felix and Næsset, Erik and Naidoo, Laven and Nilus, Reuben and O’Brien, Michael and Orwig, David A. and Papathanassiou, Konstantinos and Parker, Geoffrey and Philipson, Christopher and Phillips, Oliver L. and Pisek, Jan and Poulsen, John R. and Pretzsch, Hans and Rüdiger, Christoph and Saatchi, Sassan and Sanchez-Azofeifa, Arturo and Sanchez-Lopez, Nuria and Scholes, Robert and Silva, Carlos A. and Simard, Marc and Skidmore, Andrew and Stereńczak, Krzysztof and Tanase, Mihai and Torresan, Chiara and Valbuena, Ruben and Verbeeck, Hans and Vrska, Tomas and Wessels, Konrad and White, Joanne C. and White, Lee J. T. and Zahabu, Eliakimu and Zgraggen, Carlo},
	month = mar,
	year = {2022},
	pages = {112845},
}

@article{tolan-2024,
	title = {Very high resolution canopy height maps from {RGB} imagery using self-supervised vision transformer and convolutional decoder trained on aerial lidar},
	volume = {300},
	issn = {0034-4257},
	url = {https://www.sciencedirect.com/science/article/pii/S003442572300439X},
	doi = {10.1016/j.rse.2023.113888},
	urldate = {2024-04-30},
	journal = {Remote Sensing of Environment},
	author = {Tolan, Jamie and Yang, Hung-I and Nosarzewski, Benjamin and Couairon, Guillaume and Vo, Huy V. and Brandt, John and Spore, Justine and Majumdar, Sayantan and Haziza, Daniel and Vamaraju, Janaki and Moutakanni, Theo and Bojanowski, Piotr and Johns, Tracy and White, Brian and Tiecke, Tobias and Couprie, Camille},
	month = jan,
	year = {2024},
	pages = {113888},
}

@article{rollins-2009,
	title = {{LANDFIRE}: a nationally consistent vegetation, wildland fire, and fuel assessment},
	volume = {18},
	issn = {1049-8001, 1448-5516},
	shorttitle = {{LANDFIRE}},
	url = {https://connectsci.au/wf/article/18/3/235/23183/LANDFIRE-a-nationally-consistent-vegetation},
	doi = {10.1071/WF08088},
	language = {en},
	number = {3},
	urldate = {2026-04-14},
	journal = {International Journal of Wildland Fire},
	author = {Rollins, Matthew G.},
	month = may,
	year = {2009},
	pages = {235--249},
}

@article{safford-2026,
	title = {A collaborative, cloud-based decision support system for structured wildfire risk mitigation planning},
	volume = {514},
	issn = {0304-3800},
	url = {https://www.sciencedirect.com/science/article/pii/S0304380025004508},
	doi = {10.1016/j.ecolmodel.2025.111464},
	urldate = {2026-02-17},
	journal = {Ecological Modelling},
	author = {Safford, Hugh and Miller, Colton and Perrot, Danielle and Gilbert, Sophie and Hoecker, Tyler and Koontz, Michael and Kornhauser, Kailey and Thompson, Matt and Shannon, Joe and Rutenbeck, Nathan and Scott, Joe and Conway, Scott and Duffy, Katharyn},
	month = apr,
	year = {2026},
	pages = {111464},
}

@article{menlove-2020,
	title = {A comprehensive forest biomass dataset for the {USA} allows customized validation of remotely sensed biomass estimates},
	volume = {12},
	url = {https://research.fs.usda.gov/treesearch/63383},
	doi = {10.3390/rs12244141},
	language = {en},
	urldate = {2025-11-24},
	journal = {Remote Sensing},
	author = {Menlove, James and Healey, Sean P.},
	year = {2020},
	pages = {4141},
}

@article{chang-2025,
	title = {{VibrantVS}: {A} {High}-{Resolution} {Vision} {Transformer} for {Forest} {Canopy} {Height} {Estimation}},
	volume = {17},
	copyright = {http://creativecommons.org/licenses/by/3.0/},
	issn = {2072-4292},
	shorttitle = {{VibrantVS}},
	url = {https://www.mdpi.com/2072-4292/17/6/1017},
	doi = {10.3390/rs17061017},
	language = {en},
	number = {6},
	urldate = {2025-04-04},
	journal = {Remote Sensing},
	publisher = {Multidisciplinary Digital Publishing Institute},
	author = {Chang, Tony and Ndegwa, Kiarie and Gros, Andreas and Landau, Vincent A. and Zachmann, Luke J. and State, Bogdan and Gritts, Mitchell A. and Miller, Colton W. and Rutenbeck, Nathan E. and Conway, Scott and Bayes, Guy},
	month = jan,
	year = {2025},
	note = {Number: 6},
	pages = {1017},
}

@article{gray-2021,
	title = {Predicting canopy cover of diverse forest types from individual tree measurements},
	volume = {501},
	issn = {03781127},
	url = {https://linkinghub.elsevier.com/retrieve/pii/S0378112721007726},
	doi = {10.1016/j.foreco.2021.119682},
	language = {en},
	urldate = {2025-03-25},
	journal = {Forest Ecology and Management},
	author = {Gray, Andrew N. and McIntosh, Anne C.S. and Garman, Steven L. and Shettles, Michael A.},
	month = dec,
	year = {2021},
	pages = {119682},
}

@article{gray-2012,
	title = {Forest {Inventory} and {Analysis} {Database} of the {United} {States} of {America} ({FIA})},
	url = {https://research.fs.usda.gov/treesearch/42183},
	doi = {10.7809/b-e.00079},
	language = {en},
	urldate = {2025-01-16},
	journal = {In: Dengler, J.; Oldeland, J.; Jansen, F.; Chytry, M.; Ewald, J., Finckh, M.; Glockler, F.; Lopez-Gonzalez, G.; Peet, R. K.; Schaminee, J .H. J., eds. Vegetation databases for the 21st century. Biodiversity and Ecology. 4: 225-231.},
	author = {Gray, Andrew N. and Brandeis, Thomas J. and Shaw, John D. and McWilliams, William H. and Miles, Patrick},
	year = {2012},
	pages = {225--231},
}

@article{kennedy-2018,
	title = {An empirical, integrated forest biomass monitoring system},
	volume = {13},
	issn = {1748-9326},
	url = {https://dx.doi.org/10.1088/1748-9326/aa9d9e},
	doi = {10.1088/1748-9326/aa9d9e},
	language = {en},
	number = {2},
	urldate = {2024-08-13},
	journal = {Environmental Research Letters},
	publisher = {IOP Publishing},
	author = {Kennedy, Robert E. and Ohmann, Janet and Gregory, Matt and Roberts, Heather and Yang, Zhiqiang and Bell, David M. and Kane, Van and Hughes, M. Joseph and Cohen, Warren B. and Powell, Scott and Neeti, Neeti and Larrue, Tara and Hooper, Sam and Kane, Jonathan and Miller, David L. and Perkins, James and Braaten, Justin and Seidl, Rupert},
	month = feb,
	year = {2018},
	pages = {025004},
}

@article{riley-2021,
	title = {{TreeMap}, a tree-level model of conterminous {US} forests circa 2014 produced by imputation of {FIA} plot data},
	volume = {8},
	copyright = {2021 This is a U.S. government work and not under copyright protection in the U.S.; foreign copyright protection may apply},
	issn = {2052-4463},
	url = {https://www.nature.com/articles/s41597-020-00782-x},
	doi = {10.1038/s41597-020-00782-x},
	language = {en},
	number = {1},
	urldate = {2021-04-01},
	journal = {Scientific Data},
	publisher = {Nature Publishing Group},
	author = {Riley, Karin L. and Grenfell, Isaac C. and Finney, Mark A. and Wiener, Jason M.},
	month = jan,
	year = {2021},
	note = {Number: 1},
	pages = {11},
}

\end{document}